\documentclass[runningheads]{llncs}
\usepackage{graphicx}
\usepackage{amsmath,amssymb} 
\usepackage{color}
\usepackage{epsfig}
\usepackage{amsmath}
\usepackage{amssymb}
\usepackage{pdfpages}
\usepackage{subcaption}
\usepackage[table,xcdraw]{xcolor}
\usepackage{tabularx} 
\usepackage{commath}
\usepackage{placeins}
\usepackage[width=122mm,left=12mm,paperwidth=146mm,height=193mm,top=12mm,paperheight=217mm]{geometry}
\newcommand{\vect}[1]{{\bf {#1}}}
\newcommand{\matx}[1]{{\cal {#1}}}
\newcommand{\expect}[1]{{\mathbb{E}}{{\left[{{#1}}\right]}}}
\DeclareMathOperator*{\argmin}{argmin}
\begin{document}
\pagestyle{headings}
\mainmatter
\def\ECCV18SubNumber{1861}  
\def\httilde{\mbox{\tt\raisebox{-.5ex}{\symbol{126}}}}

\newcommand{\sign}[1]{{\mbox{sign}}{{#1}}}
\newcommand{\ea}[0]{{\em et al. }}
\newcommand{\plan}[1] {{\bf Plan:}\\ {\em {#1}}}
\newcommand{\dafkld}[2]{{\mathbb{D}({{{#1}}}\! \mid \! \mid {{{#2}}})}}
\newcommand{\pr}[1]{{\rm Pr}[#1]}
\renewcommand{\refname}{}
\renewcommand{\contentsname}{\vspace*{-1.2cm}}
\newcommand{\newterm}[1]{{\bf #1}}
\newcommand{\capsty}[1]{{\em {#1}}}
\newcommand{\argmax}[1]{{\begin{array}{c}\mbox{argmax}\\{{#1}}\end{array}}}
\newcommand{\todo}[1]{{{\bf TODO:} {{#1}}}}
\newcommand{\covmat}[1]{{
{\small {\mathsf{Covmat}}
 \left( \left\{ 
{{#1}}
      \right\} \right)
}}}
\newcommand{\mean}[1]{{{\mathsf{mean}}\left(
      \left\{{#1}\right\}\right)}}
\title{Quantitative Evaluation of Style Transfer}
\vspace{-3mm}
\authorrunning{Quantitative Evaluation of Style Transfer}

\institute{University of Illinois at Urbana and Champaign}

\author{Mao-Chuang Yeh \and Shuai Tang \and Anand Bhattad \and D. A. Forsyth\\
University of Illinois at Urbana Champaign\\
\{myeh2, stang30, bhattad2, daf\}@illinois.edu}
\maketitle
\begin{abstract}
Style transfer methods produce a transferred image which is a rendering of a content image in the manner of a style image.
There is a rich literature of variant methods.  However, evaluation procedures are qualitative, mostly involving user studies.
We describe a novel quantitative evaluation procedure.  One plots effectiveness (a measure of the extent to which the
style was transferred) against coherence (a measure of the extent to which the transferred image decomposes into objects in the same way that the content image does) to obtain an EC plot.

We construct EC plots comparing a number of recent style transfer methods.  Most methods control within-layer gram
matrices, but we also investigate a method that controls cross-layer gram matrices.
These EC plots reveal a number of intriguing properties of recent
style transfer methods.  The style used has a strong effect on the outcome, for all methods.  Using large style weights
does not necessarily improve effectiveness, and can produce worse results.  Cross-layer gram matrices easily beat
all other methods, but some styles remain difficult for all methods.  Ensemble methods show real promise.  It is likely that, for
current methods, each style requires a different choice of weights to obtain the best results, so that automated weight
setting methods are desirable.  Finally, we show evidence comparing our EC evaluations to human evaluations.
\end{abstract}
\section{Introduction}
Style transfer methods apply the {\em style} from one example image to the {\em content} of another; 
for instance, one might render a camera image (the content) as a watercolor painting (the style). 
Recent work has shown that highly effective style transfer can be achieved by searching for an image such 
that early layers of CNN representation match the early layers of the style image and later layers match the later 
layers of a content image~\cite{gatys2016image}. Content matching is by comparing activations at each location of a 
feature map. Style matching is achieved by comparing summary statistics -- in particular, the gram matrix -- of the 
layers individually. Comparing gram matrices of individual layers ensures that small, medium and large patterns that 
are common in the style image appear with about the same frequency in the synthesized image, and that spatial co-occurrences 
between these patterns are about the same in synthesized and style image.

The current evaluation of style transfer methods are done primarily by visual inspection on a small set of different styles and content image pairs. To our knowledge, there are no quantitative protocols to evaluate the competence of 
style transfer apart from user studies ~\cite{li2018closed}. This may be due to the fact that {\em styles} are subjective and more subtle to define than textures, 
hence such effectiveness metric is hard to choose. Furthermore, quick adjustment to a method using user studies is difficult in practice.
 The quantitative evaluation such as the edge coherence between {\em contents} and stylized images is investigated in ~\cite{li2018closed}. Novak and Nikulin noticed that cross-layer 
 gram matrices reliably produce improvement on style transfer (\cite{novak2016improving}). However, 
 their work was an exploration of variants of style transfer rather than a thorough study to gain insights on 
 style summary statistics. Their primary suggestions are adding more layers for more features, and they don't pursue cross-layer gram matrices 
  and quantitatively compare variant modifications. 
  
  In this paper, 
we offer a comprehensive quantitative evaluation procedure for style transfer methods. We evaluate style transfers on two criteria.  {\bf Effectiveness} measures whether transferred images have the desired style, using divergence between
convolutional feature layer distributions of the synthesized image and original image.  {\bf Coherence} measures whether the synthesized images respect the underlying
decomposition of the content image into objects, using established procedures together with the Berkeley segmentation dataset BSDS500 \cite{arbelaez2011contour}, and also using a novel measure of segment divergence.

We use our measures to compare several style transfer methods quantitatively.  
In particular, we show that controlling cross-layer, rather than within-layer, gram matrices produces quantitative improvements in style transfer over the original method due to instability in Gatys \ea proposed method (henceforth  Gatys)~\cite{gatys2016image} as described in Risser \ea~\cite{risser2017stable}.  
We construct explicit models of the symmetry groups for Gatys' style loss and the cross-layer style loss 
(improving over Risser \ea, who could not construct the groups). We discuss this in detail in section \ref{sec:symmetry}.  We show experimental evidence that 
the quantitative improvement over Gatys' method is due to the difference in symmetry groups. 
We show qualitative evidence suggesting that these quantitative improvements manifest in real images.

\section{Related work} \label{sec:gatys}
Bilinear models are capable of simple image style transfer~\cite{Tenenbaum2000} by factorizing style and content representations, but non-parametric methods like patch-based texture synthesis can deal with much more complex texture fields~\cite{Efros2001}.  Image analogies
use a rendering of one image in two styles to infer a mapping from a content image to a stylized
image~\cite{Hertzmann2001}. Researchers have been looking for versatile parametric methods to control style patterns at
different scales to be transferred. Adjusting filter statistics is known to yield texture
synthesis~\cite{debonet,simoncelli}.  Gatys \ea demonstrated that producing neural network layers with particular
summary statistics (i.e Gram matrices) yielded effective texture synthesis~\cite{NIPS2015_5633}. In a following paper,
Gatys \ea achieved style transfer by searching for an image that satisfies both style texture summary statistics and
content constraints~\cite{gatys2016image}. This work has been much elaborated. The search can be replaced with a
regression (at one scale~\cite{Johnson2016Perceptual}; at multiple scales~\cite{wang2016multimodal}; with
cached~\cite{chen2017stylebank} or learned~\cite{dumoulin2016learned} style representations) or a decoding process that
allows efficient adjusting of statistics ~\cite{DBLP:journals/corr/UlyanovVL16,huang2017arbitrary,UST,li2018closed}. Search can be sped up with local matching
methods~\cite{chen2016fast}. Methods that produce local maps (rather than pixels) result in photorealistic style
transfer~\cite{Shih2014,Luan2017}. Style transfer can be localized to masked regions~\cite{gatys2016controlling}. The
criterion of matching summary statistics is a Maximum Mean Discrepancy condition~\cite{li2017demystifying}. Style
transfer has also been used to enhance sketches~\cite{champandard2016semantic}.There is a comprehensive review in~\cite{jing2017neural}.

Gupta \ea ~\cite{gupta2017characterizing} study instability in style losses from videos, 
where they use prior video frames to stabilize current video frame by enforcing a temporal consistency loss. They demonstrate theoretically
instability in Gaty's method is linked to the size of the trace of the gram matrix. They support this argument with experimental evidence
that larger traces result in higher instability.

\vspace{-2mm}
\subsection{Gatys Method}
We review the original work of Gatys \ea \cite{gatys2016image} in detail to introduce notation.
Gatys finds an image where early layers of a CNN representation match the lower layers of the style image and higher layers match the higher layers of a content image.  Write $I_{s}$ (resp. $I_{c}$, $I_{n}$)  for the style (resp. content, new) image,
and $\alpha$ for some parameter balancing style and content losses ($L_s$ and $L_c$ respectively).  Occasionally, we
will write $I_n^m(I_c, I_s)$ for the image resulting from style transfer using method $m$ applied to the arguments.
We obtain $I_{n}$ by finding

\[
{\argmin_{I_n}} L_c(I_{n}, I_{c})+\alpha L_s(I_{n}, I_{s})
\]

Losses are computed on a network representation, with $L$ convolutional layers, where the $l$'th layer
produces a feature map $f^l$ of size $H^l \times W^l \times C^l$ (resp. height, width, and channel number).  We partition
the layers into three groups (style, content and target). Then we reindex the spatial variables (height and width) and
write $f^l_{k,p}$ for the response of the $k$'th channel at the  $p$'th location in the $l$'th convolutional layer. The
content loss $L_c$ is 

\[
L_c(I_{n}, I_{c}) = \frac{1}{2}\sum_{c} \sum_{k,p} \norm{f^c_{k,p}(I_{n}) - f^c_{k,p}(I_{c})}^2
\]

\noindent (where $c$ ranges over content layers). The {\em within-layer gram
  matrix} for the $l$'th layer is
\[
G_{ij}^l(I) = \sum_p \left[f_{i,p}^l(I)\right]\left[f_{j,p}^l(I)\right]^{T}.
\]

\noindent Write $w_l$ for the weight applied to the $l$'th layer.  Then 
\[
L_s^l(I_{n}, I_{s}) = \frac{1}{4{N^l}^2{M^l}^2}\sum_{s}w_l \sum_{i,j}\norm{G^s_{ij}(I_{n})-G^s_{ij}(I_{s})}^2
\]

\noindent where $s$ ranges over style layers. Gatys \ea use Relu1\_1, Relu2\_1, Relu3\_1, Relu4\_1, and Relu5\_1 as style layers, and layer Relu4\_2 
for the content loss, and search for $I_{n}$ using L-BFGS~\cite{liu1989limited}.  From now on, we write R51 for Relu5\_1, etc. 

\section{Quantitative Evaluation of Style Transfer}\label{effcoh}

A style transfer method should meet two basic tests.  The first is {\bf effectiveness} -- does the
method produce images in the desired style? The second is {\bf coherence} -- do the resulting images respect the
underlying decomposition of the content image into objects?   While final judgment should belong to the artist, we
construct numerical proxies that can be used to disqualify methods from a final user study. 
It is essential to test both properties (excellent results on coherence can
be obtained by simply not transferring style at all).  In this paper, we offer one possible
effectiveness statistic and two possible coherence statistics; however, we expect
other reasonable choices could apply.

{\bf Effectiveness:}  Assume that a style is applied to a content image. We would like to measure the extent to which the result reflects the style.  
There is good evidence that the distribution of features within lower feature layers of a CNN 
representation is an effective proxy to capture styles ~\cite{bau2017network}.  
We expect that individual transferred images might need to have small biases in the distribution of 
feature layers to account for the content, but over many images the distribution of features 
should reflect the style distribution.  In turn, a strong measure of effectiveness of style 
transfer for a particular image is the extent to which the distribution of feature layer values 
produced by the transferred image matches the corresponding distribution for the style image. 
In notation, write $\vect{f}^{l}_{p}(I)$ for the vector of responses  of all channels  at the  
$p$'th location in the $l$'th convolutional layer for image $I$. Now choose the $i$'th content 
image, the $j$'th style, and some method $m$.  
The distribution $P_{t, m}$ of $\vect{f}^{l}(I^m_n(I_c^i,I_s^j))$ should be similar to  
the distribution $P_s$ of $\vect{f}^{l}(I_s^j)$, with perhaps some smoothing  resulting 
from the need to meet content demands.   

Testing whether two datasets come from the same, unknown, distribution in high dimensions remains tricky 
(the method of~\cite{gretton2012kernel} is the current best alternative).  We do not expect the distributions
 to be exactly the same;  instead, we want to identify obvious (and so suspicious) large  differences. 
 The symmetry analysis below suggests that Gatys method will massively increase the variance of  
 $\vect{f}^{l}(I^g_n(I_c^i,I_s^j))$.  Observing major differences is straightforward with relatively 
 crude tools.   However, dimension is a problem.  Even assuming that each distribution is normal,  
 computing KL divergences is impractical, because the distributions are large and so the estimates 
 of the covariance matrices are unreliable.   

However, we seek a statistic that is large with high probability when $P_{t,m}$ and $P_s$ are strongly 
different, and small with high probability when they are similar.  A straightforward construction, 
after~\cite{DeshpandeCVPR2018} is as follows .
Write $\vect{v}_k$ for a random unit vector.  We then compute
$p_p^m=\vect{v}_k^T\vect{f}_p^{l}(I^m_n(I_c^i,I_s^j))$ and $p_p^s=\vect{v}_k^T\vect{f}_p^{l}(I_s^j)$.  We assume that
these scalar datasets are normally distributed, and compute KL divergence $d(\vect{v}_k)$ from the style distribution to the transferred
distribution.  We now average over $R$ random unit vectors and form
\[
E=-\log \left(\frac{1}{R} \sum_{k}d(\vect{v}_k)\right)
                     \]                   
Large values of this statistic are obtained if there are few random directions in which the two distributions 
differ; small values suggest there are many such directions and so that the style transfer may not 
have succeeded. For all our analysis, we choose a single set of 128 random unit vectors that is reused for all methods.

{\bf Coherence:} A style transfer method that eliminates object boundaries would make it hard for humans to interpret
the output images, so a reasonable measure of a style transfer method is the extent to which it preserves object
boundaries.  We have two measures of coherence.  Our {\em boundary preservation} measure computes the extent to which a
boundary prediction algorithm produces true object boundaries for a given method, using the Berkeley segmentation dataset tests BSDS500~\cite{arbelaez2011contour}.  Our {\em object coherence} measure computes the extent to which textures are (a) coherent within object boundaries and (b) distinct from object to object.
{\em Boundary preservation} is treated as a straightforward application of existing methods to evaluate image boundaries.  We choose a boundary predictor (we used the contour detection of~\cite{arbelaez2011contour}); we apply the style transfer methods to images
from the BSDS500, using multiple style images, to obtain synthesized images; we apply the boundary predictor to the synthesized images; and we compute the area under curve (AUC) of the  probability of boundary (Pb) precision-recall curve for every synthesized image.  A higher AUC suggests better boundary
preservation.  As section~\ref{results} shows, this measure is highly variable depending on the style that is transferred, and so we compute a per-transferred image AUC. This evaluation method is not perfect.  Heavily textured styles may confuse the Pb evaluation without confusing human viewers, because the contour detector was not built with very aggressive texture fields in  mind (compare typical style transfer images with the  ``natural'' textures used to build BSDS500).  In particular, we might have texture fields that are strongly coherent within each object region and different from region to region, but where the contour detector has great difficulty identifying object boundaries.

An \textit{object coherence} measure is easy to obtain using the BSDS500 dataset, because each image comes with a ground truth contour
mask.  We choose some layer $l$, and write $\vect{f}_{S,i}=\vect{f}_{i}^{l}(I^g_n(I_c^i,I_s^j), S)$ (for brevity) for a feature
vector in that layer within some segment $S$, and $\left\{\vect{f}_{S,i}\right\}$ for all such feature
vectors.  Write $\mu_S=\mean{\vect{f}_S}$, and $\Sigma_b=\covmat{\mu_S}$ for the between class covariance matrix.
Assume that each segment has the same covariance(heteroskedasticity,
a tolerable assumption given that the method tries to impose a gram matrix on the layer), and construct the within-class covariance for all locations in a segment 
$\Sigma_w=\covmat{\vect{f}_{1,1}-\mu_1, \ldots \vect{f}_{n_S,n_f(S)}-\mu_{n_S}}$.
Now the
largest generalized eigenvalue $\lambda^{\mbox{max}}$ of $(\Sigma_b, \Sigma_w)$ measures the dispersion of the region
textures.   Notice that $\lambda^{\mbox{max}}\geq 0$, and simple plots (supplementary materials) suggest this has a
log-normal distribution over multiple style/content pairs.  We therefore use $L_m=\log \lambda^{\mbox{max}}_{m}$ as a
score to evaluate a method.  Larger values suggest more successful separation of regions.\\

{\bf Summarizing data with the EC plot.} Comparing style transfer methods requires a 
summary of: the expected effectiveness of a method at any coherence; the effect of style
 and of weight choice on performance; and the extent to which evidence supports a difference between methods.
 We compare methods using an effectiveness-coherence (EC) plot, which plots: 
 (a) a scatter plot of EC pairs obtained for various style/content/weight triplets;   
 (b) a Loess regression curve of E regressed against C for these triplets; and 
 (c) standard error regions for the regression.     
 Effectiveness is measured per layer and we show layer 1 plots in section \ref{ExpPro} (with others in the supplementary material). Coherence is measured either using per-image AUC of Pb (which does not depend on layers) or using $L_m$, \textit{object coherence}; this depends on the layer (more plots in supplementary material).
\begin{figure*}[!t]
    \centering
    {\includegraphics[clip, trim=1cm 8cm 1cm 3cm, width=\linewidth]{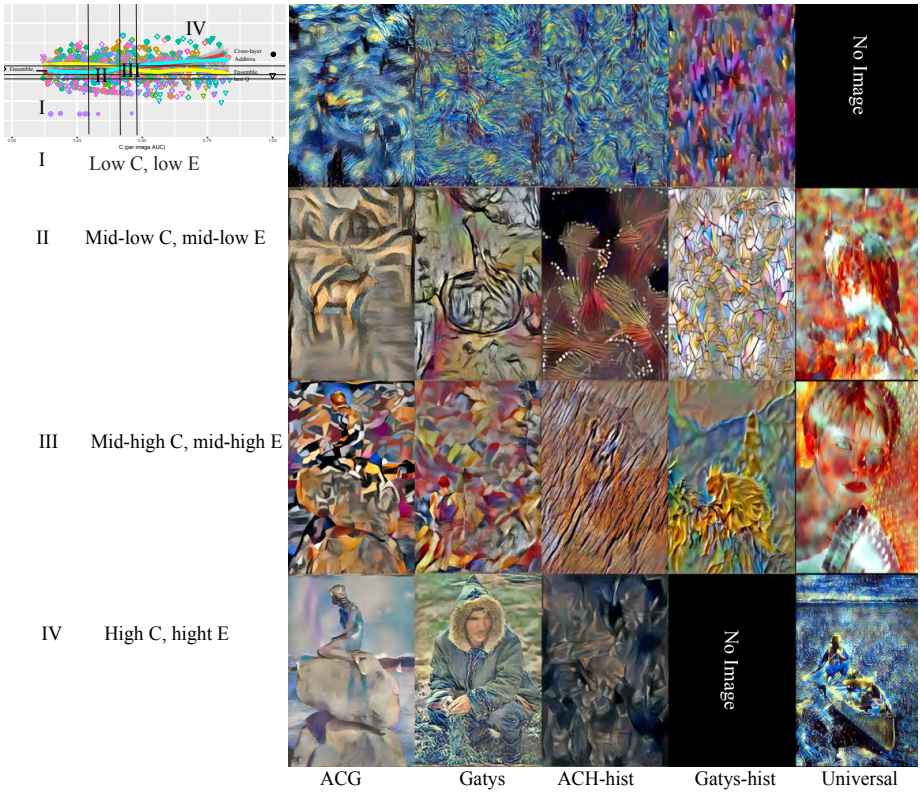}}
    \caption{To visualize EC space, we divide it into a 4x4 grid of boxes using quantiles of all E (resp. C) values for
all methods.  This grid is shown in the inset; notice boxes II and III are quite small, meaning that there are
many images with only moderate scores over all the methods.  We then obtain for each method and for each anti-diagonal
box the mediod of that methods images within the corresponding box; in some cases, there are no images for
that method in the relevant box.  We have laid these images out as a map of EC space.  Notice that the space is
reasonably consistent with intuition; images that achieve strong E and strong C are both stylish and recognizably composed
of objects; images that achieve weak E and weak C have rather poor style and do not noticeably contain objects. Note that ``No-Image'' indicates here there are no stylized image in the quantiles in portrait content; not necessary to have zero samples in landscape mode (see sec. ~\ref{ExpPro} and refer supplementary figures). Best
viewed at high resolution in color.}  
    \label{fig:qual_port}
    \vspace{-2mm}
    \end{figure*} 
\section{Cross-layer Style Transfer}\label{sec:Cross}

\subsection{Cross-layer style loss}
We consider a style loss that takes into account between layer statistics.  The {\bf cross-layer, additive (ACG)} loss
is obtained as follows.  Consider layer $l$ and $m$, both style layers, with decreasing 
spatial resolution.  

Write $\uparrow f^{m}$ for an upsampling of  $f^m$ to $H^l\times W^l \times K^m$, and consider
\[
G_{ij}^{l,m}(I) = \sum_{p} \left[ f_{i,p}^l(I)\right]\left[\uparrow {f}_{j,p}^{m}(I)\right]^{T}.
\]
as the cross-layer gram matrix, We can form a style loss
\[
L_s(I, I_{s}) = \sum_{(l, m)\in {\cal L}} w^{l}\sum_{ij} \norm{G^{l,m}_{ij}(I)-G^{l,m}_{ij}(I_s)}^2
\]
(where ${\cal L}$ is a set of pairs of style layers).   We can substitute this loss into the original style loss, and
minimize as before.  All results here used a {\em pairwise descending} strategy, where one constrains each layer and its
successor (i.e. (R51, R41); (R41, R31); etc).  Alternatives include an {\em all distinct pairs} strategy, where one constrains all pairs of distinct
layers. Carefully controlling weights for each layer's style loss is not necessary in cross-layer gram matrix scenario.  
{\em Constraint counts} for cross-layer gram matrix methods are much lower than for with-in layer methods.  For a
pairwise descending strategy,  we have four cross-layer gram matrices, leading to  control of $64\times
128+128\times 256+256\times 512+512\times 512 = 434176 $ parameters; compare within layer gram matrices, which control 
$64^2+128^2+256^2+2\times512^2 = 610304$ parameters.  The experimental results suggest that the number of constraints is a poor way of evaluating a method.

\vspace{-2mm}
\subsection{Symmetries and Stability}\label{sec:symmetry}
 
Symmetries in a style transfer loss function occur when there is a transformation available that changes the style
transferred image without changing the value of the loss function.   Risser \ea note instability in Gatys' method;
symptoms are poor and good style transfers of the same style to the same content with about the same loss value~\cite{risser2017stable}.
They supply evidence that this behavior can be controlled by adding a histogram loss, which breaks the symmetry.
They do not write out the symmetry group as too complicated
(~\cite{risser2017stable}, p 4-6).   Gupta \ea ~\cite{gupta2017characterizing} make a strong experimental argument that instability in Gaty's method is linked to the size of the trace of the gram matrix (larger trace is linked to more instability).

One portion of the symmetry group is easy to construct.  In particular, we consider affine maps acting on a feature layer, and consider the effect on that layers gram matrix and
on the gram matrix of the next layer.  Notice this does not exhaust the available symmetries (for example, a spatial permutation of features
would not change the gram matrix).  We have no construction currently for spatial symmetries.
The supplementary materials give a construction for all affine maps that fix the gram matrix for a layer and its parent
(deeper networks follow the same lines).  It is necessary to assume the map from layer to layer is linear.  This is not
as restrictive as it may seem; the analysis yields a local construction about any generic operating point of the
network.  In summary, we have:

{\bf Symmetry group, within layer gram matrices, two layers:}
Assuming that the between layer map is affine, with matrix $\matx{M}$
representing the linear component.  With various assumptions about the
spatial statistics of layer 1 (supplementary materials), an element of the symmetry
group is obtained by:  choose $\vect{b}$ {\em not} of unit length, and
such that $\matx{M}\vect{b}=0$; now factor
$\matx{I}-\vect{b}\vect{b}^T=\matx{A}\matx{A}^T$; choose $\matx{U}$
orthonormal.  Then $(\vect{b}, \matx{A}\matx{U})$ is a symmetry of the
gram matrices in {\em both layers} (i.e the action of this element on
layer 1 fixes {\em both} gram matrices).   In particular, mapping all feature vectors $\vect{f}_p^1$ to
$\matx{A}\matx{U}\vect{f}_p^1+\vect{b}$ will result in no change in the gram matrix at either layer 1 or layer 2; but
the underlying image may change a lot, because $\matx{A}$ can rescale features and features are shifted.

{\bf Symmetry group, between layer gram matrix, two layers:}
Assuming that the between layer map is affine, with matrix $\matx{M}$
representing the linear component.  With various assumptions about the
spatial statistics of layer 1 (supplementary materials), the symmetry
group is obtained by:  choose $\matx{U}$
orthonormal.  Then $(\matx{U})$ is a symmetry of the between layer
gram matrix (i.e the action of this element on layer 1 fixes the
between layer gram matrix).  In particular, mapping all feature vectors $\vect{f}_p^1$ to
$\matx{U}\vect{f}_p^1$ will result in no change in the gram matrix at either layer 1 or layer 2; we expect much less
change in the underlying image.

The between-layer gram matrix loss has very different symmetries to Gatys'
(within-layer) method.  In particular, the symmetry of Gatys' method rescales
features while shifting the mean (because in this case $\matx{A}$ can contain
strong rescalings with the right choice of $\vect{b}$).   For the
cross-layer loss, the symmetry cannot rescale, and cannot shift the
mean.   This implies that, if one constructs numerous style
transfers with the same style using Gatys' method, the variance of the layer features should be much greater than that
observed for the between layer method. 

Furthermore, increasing style weights in Gatys method should result in poor
style transfers, by exaggerating the effects of the symmetry.    Finally, our construction casts light on part Gupta \ea's
observation linking large trace to instability. A small trace in the gram matrix implies many small eigenvalues.  In
turn, rescaling directions with small eigenvalues will change little unless very large scales are applied; but these
correspond to very large shifts in the mean, which are difficult to obtain with current random start methods.  However,
a large trace in the gram matrix implies that there are many directions where a small shift in the mean will result in a
small -- but visible, because the eigenvalue is big -- rescale from $\matx{A}$ will lead to real changes, and so there
is greater instability. 

Note that this analysis is limited by the fact that strong scales and shifts will likely cause RELU's to change state,
by the fact that it takes no account of the content loss, and by the absence of spatial symmetries.
But the analysis exposes the fact that quite large changes in early layers will leave the style loss unchanged.
Since we expect that at least some large changes in early layers will produce very little change in content
layers (otherwise image classification applications would not work), the analysis is a fair rough guide. Experimental
observations are consistent with the symmetry theory  (figure~\ref{aggweight}; and
section~\ref{results}).

\section{Experimental Procedures\label{ExpPro}}

{\bf Comparison data:}  It is important to do comparisons on a wide range of styles and contents.  
We have built two datasets, using 50 style images (see supplementary) and the 200 content images from the BSDS500  test set. The {\em main set} is used for most experiments, and was obtained by:  
take 20 evenly spaced weight values in the range 50-2000; 
then, for each weight value, choose 15 style/content pairs uniformly and at random.  
The {\em aggressive weighting set} is used to investigate the effect of extreme weights 
on Gatys method and the ACG method.  This was built by taking 20 weight values sampled uniformly and 
at random between 2000-10000; then, for each weight value, 
choose 15 style/content pairs uniformly and at random.   
For each method, we then produced 300 style transfer images using each weight-style-content triplet.  
For UST~\cite{UST}, since the maximum weight is one, 
we linearly map \textit{main set} weights to the zero-one range. 
Our samples are sufficient to produce 
clear differences in standard error bars and evaluate different methods ~\cite{Forsyth2018}. 
{\bf Methods.}  We compare the following methods:
\noindent {\em Gatys} (\cite{gatys2016image} and described above); we use the implementation by Gatys~\cite{leongatys}.
\noindent {\em ACG:} We used a {\em pairwise descending} strategy with pre-trained VGG-16 model. We use R11, R21, R31, R41, and R51 for style loss, and R42 for the content loss for style transfer.\\
\noindent {\em Cross-layer, multiplicative (MCG):}  A natural alternative to combine style and content losses is to multiply them; we form 
\[
L^m(I_n) = L_c(I_n, I_c) *  L_s(I_n, I_s).
\]
It provides a dynamical weighting between content loss and style loss during optimization. Although this loss function seems unreasonable, but we find them to perform competitively on a wide range of our EC plots (see supplementary).

\noindent {\em Gatys, with histogram loss:}  as advocated by \cite{risser2017stable}, we attach a histogram loss to
Gatys method.  
\noindent {\em ACG, with histogram loss:}  We use the implementation of ~\cite{abhiskk} for histogram adjustment.\\
\noindent {\em Universal style transfer:}(from \cite{UST}, and it's Pytorch implementation~\cite{SunshineAtNoon} 
;\\
\noindent {\em Ensemble Q:}  for each weight-style-content triple, we choose the result that produces the best  $Q=E*C$
over all methods.\\
\noindent {\em Ensemble E:} for each weight-style-content triple, we choose the result that produces the best  $E$
over all methods.

\begin{figure*}[t]
\centering
  \includegraphics[width=1\linewidth]{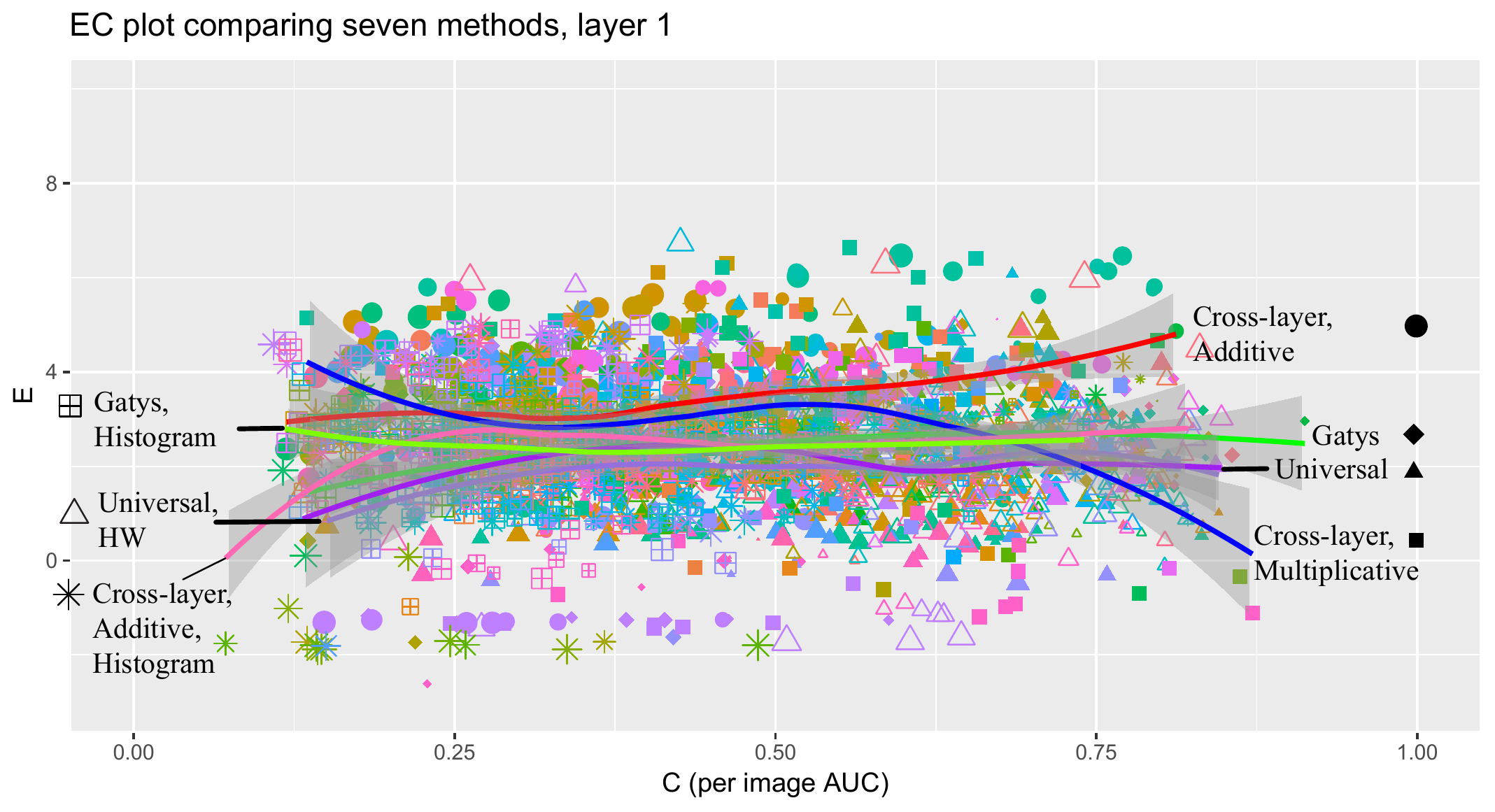}
\caption{\em An EC plot comparing style transfers methods.  Here the C
statistic is per image Pb AUC.  Point markers show individual image
statistics, with color keyed
to the style and size keyed to the weight on the style loss (larger markers corresponds to a stronger style weight).  Notice that some styles are clearly harder than others, and produce
low E for both methods.  The curves are Loess regression curves of E against C, with shadowed regions showing one
standard error bars up and down.  For any value of C, there is strong evidence that ACG obtains on
average a larger E than Gatys' loss (about three standard errors difference); 
Note that cross-layer loss achieves an E comparable with simply resizing the style image( see Fig. ~\ref{control}, style only control). A similar plot can be obtained by choosing C as segment divergence by plotting largest eigenvalues against E (plots in supplementary material). Best viewed in color.}
\label{ecxvg}
\vspace{-6mm}
\end{figure*}


\section{Results\label{results}}
\begin{figure*}[t]
\centering
  \includegraphics[width=\linewidth]{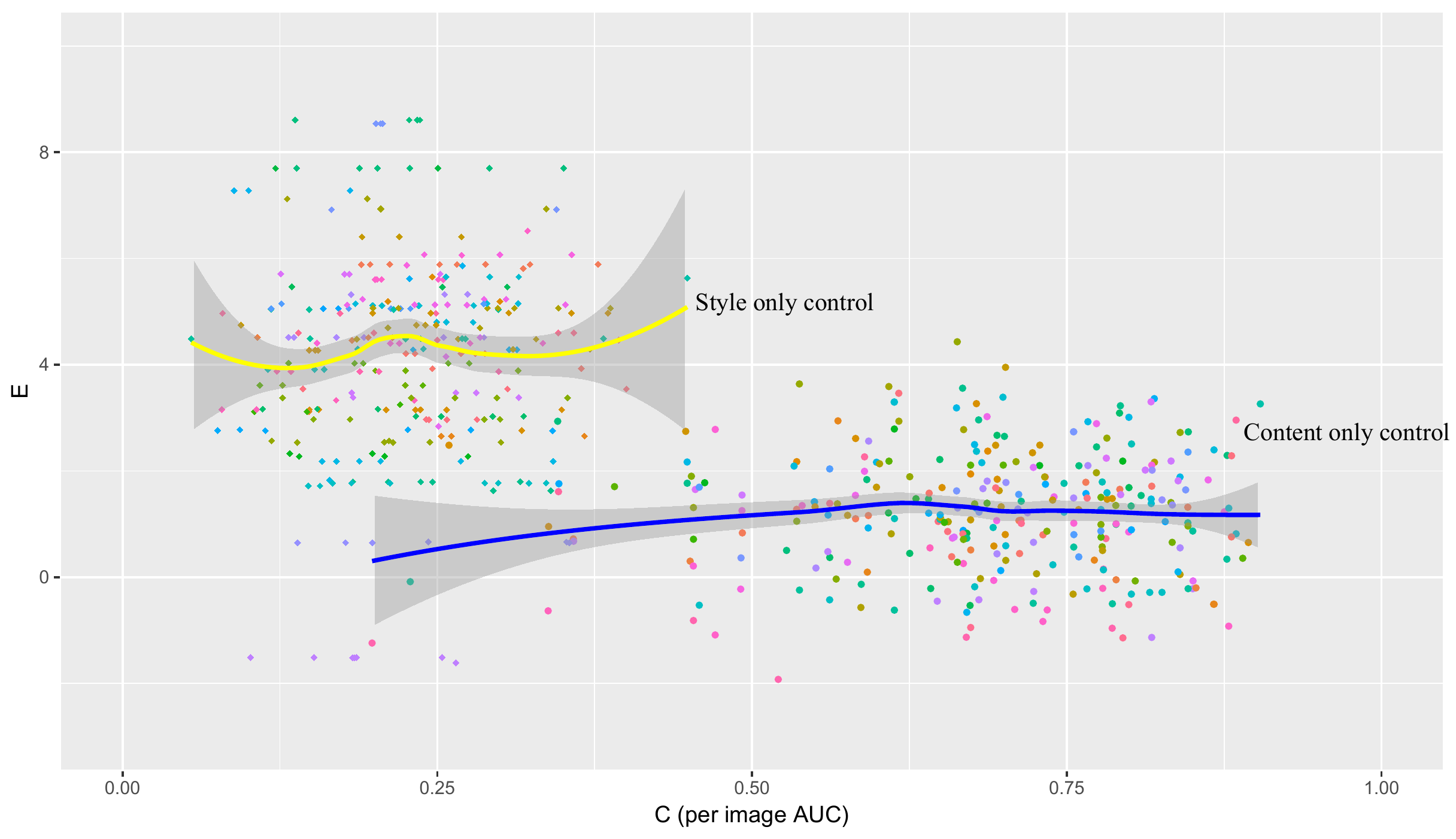}
\caption{\em An EC plot comparing two control methods.  One reports the resized style image as the transfer (small diamonds,
yellow curve) and the other reports the content image as the transfer (large dots, blue curve).  Colors are keyed to
style. The curves are Loess regression curves of E against C, with shadowed regions showing one standard error bars. Scale is the same as Figure \protect \ref{ecxvg}.     }
\label{control}
\vspace{-5mm}
\end{figure*} 


{\bf ACG is better.} Figure~\ref{ecxvg} shows an EC plot of layer 1 comparing style transfers using the cross-layer additive loss (ACG) with transfers using five other non-ensemble losses. Note that cross-layer loss achieves much higher average E for a given value of C.  In various parts of the AUC range, additive cross-layer loss is somewhat outperformed by the multiplicative version, but in the high AUC regime it is much better.  \
The difference to every other method ranges from one to four standard errors over the range, hence our 300 sample size is large enough~\cite{Forsyth2018}; the ACG method is clearly significantly better.

{\bf Control.}  Figure~\ref{control} shows an EC plot of two controls.  In the first, the resized style image is reported as a transfer; this results as expected in high values of E, but low values of C. There is significant variance in E, an effect due to resizing.  However, the range of E's shows the size of the effect of resizing on E; on average, E slightly greater than 4.  In the second, the content image is reported as a transfer; this results  suggest that obtaining high C values (though not uniformly; some images remain hard to segment)  may at the cost of getting low E values nearly to zero (look at the differences of E values for two controls). This shows investigating E is necessary for all methods. .

{\bf Histogram losses.} improves Gatys' method (compare green/light green on figure~\ref{ecxvg}) only at extreme weights and low C. This may be an effect of the loss of symmetry, explained below. They also weaken the performance of cross-layer style transfer (compare red/pink on figure~\ref{ecxvg}, about three standard errors). 

\begin{figure*}[t]
\centering
    \includegraphics[width=0.95\linewidth]{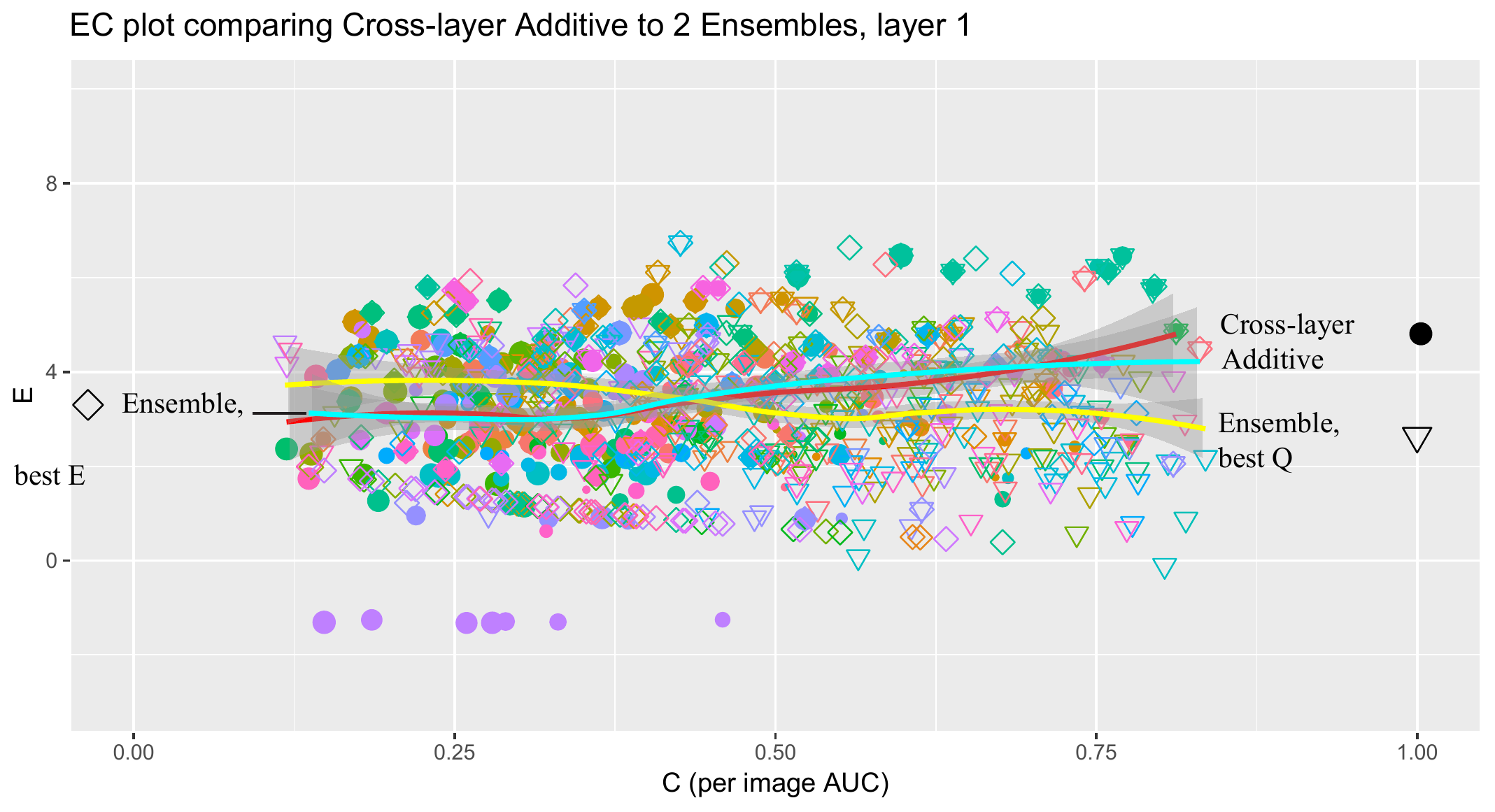}
\caption{\em An EC plot comparing two ensemble methods to ACG.  
Scale is the same as Figure \protect \ref{ecxvg}.   }
\label{ensmethl}
\vspace{-5mm}
\end{figure*}

{\bf Ensemble methods.} do not outperform cross-layer style transfer (see figure~\ref{ensmethl}).  As comparing the red and cyan curves suggests, choosing the result with the best E is essentially the same as using the cross-layer style transfer result. The yellow curve shows the ensemble Q method, which works somewhat better than cross-layer style transfer in low C regimes, and somewhat worse in high C regimes.  This suggests that more sophisticated ensemble methods might yield even better performance.

\begin{figure*}[!htbp]
\centering
    \includegraphics[width=0.95\linewidth]{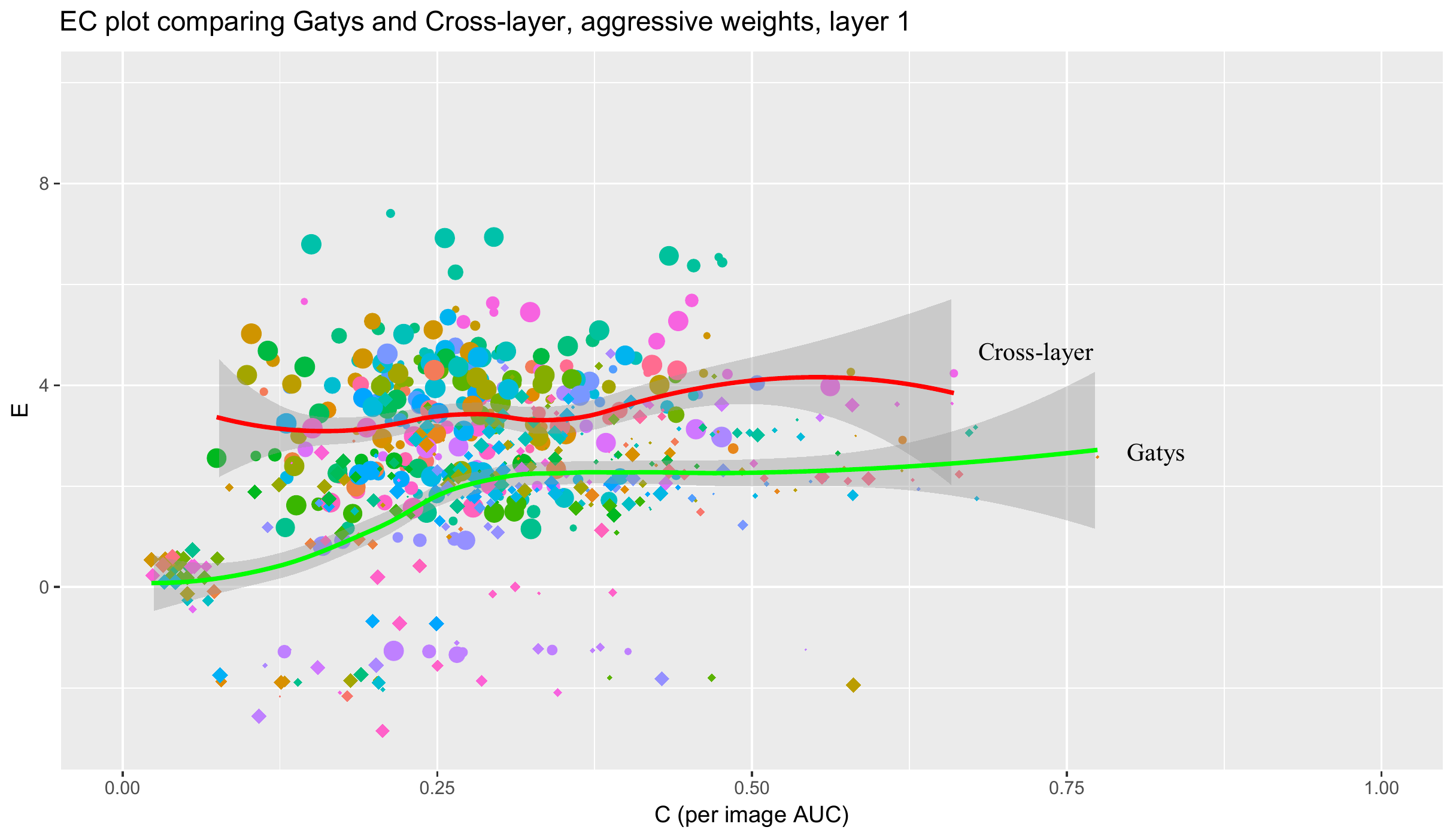}
\caption{\em An EC plot comparing Gatys' method to cross-layer style transfer for the aggressive weight dataset. Notice that large weights cause serious trouble for
Gatys' method (large diamonds clustered in the bottom left corner). This is very likely an effect of the symmetry in the loss function (text in section \ref{sec:symmetry}).
Cross-layer style transfer outperforms Gatys' method over all of its range, mostly by many
standard errors. Scale is the same as Figure \protect \ref{ecxvg}.   }
\label{aggweight}
\vspace{-7mm}
\end{figure*}

{\bf Aggressive weights.} One might speculate that Gatys' method underperforms because the weight regime is inappropriate. Figure~\ref{aggweight} compares Gatys' method to cross-layer style transfer. Notice that large weights cause serious trouble for Gatys' method.  We believe that this is because large weights on the style loss cause the symmetry in Gatys' loss to manifest itself, resulting in significant rescaling of features. In particular, Gatys' method cannot achieve high E values, because symmetry in the style loss produces feature distributions that are quite different from that desired. Furthermore, larger weights on the style loss {\em do not} produce better style transfers (large diamonds toward the bottom left of the plot).  Instead, by exaggerating the effect of the symmetry, large weights produce transfers that both have low E (poor transfer) {\em and} low C (do not respect original segmentation).  

\textbf{User Study.} We obtained preference data from Amazon Mechanical Turk (AMT) workers. We use 300 \textit{main set} image pairs from ACG and Gatys results, each image pair is annotated by 10 workers in total, separated in two groups. Detailed worker-task statistics are present in supplementary materials

Mechanical turk worker data is extremely noisy, and so difficult to plot helpfully.  We distinguish between 44 master workers (who are known to be quite reliable) and 49 generic workers (about whom we know nothing).  We analyze using a logistic regression of preference for cross-layer (resp. Gatys) against
E values for Gatys and for cross-layer and C values for Gatys and for cross-layer.  Our analysis supports the idea that E and C predict worker preferences, but that
there are other likely sources of preference, too (or that better measures of E and C would help).  We have two datasets: one using master workers only, and the other using
workers of any type. \\
 {\em Analysis, master workers:}  all four regression coefficients and the intercept are different from zero with strong statistical significance  (for each coefficient, $p<0.025$).  Weights produced by this regression are:   Intercept: 0.3409;    $E_{Gatys}= -0.1484;    E_{ACG}=  0.1015;C_{Gatys}=-3.4369;C_{ACG}=3.8982$ \\
{\em Conclusion, master workers:} master workers slightly prefer cross-layer images over Gatys images, whatever E and C;
worker preference can be predicted by looking at E and C; in particular, master workers tend to prefer transfers with higher E and C values (if cross-layer has higher E and C, it will tend to be preferred, etc).  The difference in weight size is roughly proportional to the relative scales (a factor of about 10), but one measure may be more important to workers than others.  The regression has relatively high deviance (and
cross-validated AUC of predictions by this regression is approximately 0.57, depending on regularization constant), meaning that other factors may explain preferences, too.  \\
{\em Analysis, generic workers:}  three of four regression coefficients are different from zero with strong statistical significance (for each coefficient, $p<0.02$, but the intercept could be zero).  Significant weights produced by this regression
are:$E_{ACG}=0.076, C_{Gatys}=-3.38, C_{ACG}=3.77$.  However, a cross-validated L1 regularized logistic regression obtains an average AUC on held-out predictions of about 0.85 (depending on regularization coefficient) using only the $E_{Gatys}$ and $C_{ACG}$ coefficients; this suggests that a preference for cross layer images is predicted by large values E and C on Gatys images. \\
{\em Conclusion, generic workers.} worker preference can be predicted by looking at C, and checking whether the $E_{ACG}$ is large; in particular, workers tend to prefer transfers with higher  C value (if cross-layer has higher  C, it will tend to be preferred, etc).  The effect of E is small.  The regression has relatively high deviance, and there is good evidence of odd behavior by workers (who prefer cross-layer images when E and C is larger for Gatys), meaning there may be workers who are not attending to the task.

{\bf The experimental effects of symmetry:}  Our experimental evidence suggests the symmetries manifest themselves in
practice. Gatys' method significantly underperforms the cross-layer method by producing a lower E statistic for any C
statistic.  This suggests that the variance implied by the larger symmetry group is actually appearing.  In particular,
Gatys' symmetry group allows rescaling of features and shifting of their mean, which will cause the feature distribution
of the transferred image to move away from the feature distribution of the style, causing the lower E statistic.
Furthermore, Gatys' method has a strong tendency to produce very poor transfers when offered aggressive weighting of the
style loss.  We believe this is likely because large rescaling effects are suppressed when the style loss has a smaller
weight, because large rescaling will eventually lead to a change in the content loss.  But when the style loss has a
high weight, then the changes in the content loss are of small significance, and very significant variations can appear in the
early layers, forcing down the E value; the C value goes down because little weight is placed on the content loss. This
effect does not appear for the cross layer method, because rescaling isn't possible for those symmetries.
\vspace{-2mm}
 \section{Conclusion}
 \vspace{-1mm}
 In this paper we present a novel approach to quantitatively evaluate style transfer methods. Our metric is built with two factors in mind, Effectiveness: a good style transfer should preserve the desired characteristics of original style; Coherence: style transfer method should respect to content's underlying decomposition of object segments. We apply various style transfer methods which are built either on with-in layer or cross-layer gram matrices, and we compared stylized images both quantitatively using the proposed EC plots, and qualitatively showing their results as well as conducting user study. Using this analytical framework, we confirm Gatys method is troubled by symmetry group, especially so when having aggressive style weights. The cross-layer method, which has very different symmetry group setting, is less compromised and thus achieves higher EC score. This conclusion is supported by master AMT workers' preference from user study.       


\clearpage

\section*{References}
\vspace{-5mm}
{\small
\bibliographystyle{splncs}
\bibliography{main1}

\begin{thebibliography}{10}

\bibitem{gatys2016image}
Gatys, L.A., Ecker, A.S., Bethge, M.:
\newblock Image style transfer using convolutional neural networks.
\newblock In: Proceedings of the IEEE Conference on Computer Vision and Pattern
  Recognition. (2016)  2414--2423

\bibitem{li2018closed}
Li, Y., Liu, M.Y., Li, X., Yang, M.H., Kautz, J.:
\newblock A closed-form solution to photorealistic image stylization.
\newblock arXiv preprint arXiv:1802.06474 (2018)

\bibitem{novak2016improving}
Novak, R., Nikulin, Y.:
\newblock Improving the neural algorithm of artistic style.
\newblock arXiv preprint arXiv:1605.04603 (2016)

\bibitem{arbelaez2011contour}
Arbelaez, P., Maire, M., Fowlkes, C., Malik, J.:
\newblock Contour detection and hierarchical image segmentation.
\newblock IEEE transactions on pattern analysis and machine intelligence
  \textbf{33}(5) (2011)  898--916

\bibitem{risser2017stable}
Wilmot, P., Risser, E., Barnes, C.:
\newblock Stable and controllable neural texture synthesis and style transfer
  using histogram losses.
\newblock arXiv preprint arXiv:1701.08893 (2017)

\bibitem{Tenenbaum2000}
Tenenbaum, J.B., Freeman, W.T.:
\newblock {Separating Style and Content with Bilinear Models}.
\newblock Neural Computation \textbf{12}(6) (2000)  1247--1283

\bibitem{Efros2001}
Efros, A.A., Freeman, W.T.:
\newblock {Image quilting for texture synthesis and transfer}.
\newblock Proceedings of the 28th annual conference on Computer graphics and
  interactive techniques - SIGGRAPH '01 (2001)  341--346

\bibitem{Hertzmann2001}
Hertzmann, A., Jacobs, C.E., Oliver, N., Curless, B., Salesin, D.H.:
\newblock {Image analogies}.
\newblock Proceedings of the 28th annual conference on Computer graphics and
  interactive techniques - SIGGRAPH '01 (August) (2001)  327--340

\bibitem{debonet}
Bonet, J.D.:
\newblock Multiresolution sampling procedure for analysis and synthesis of
  texture images.
\newblock SIGGRAPH (1997)

\bibitem{simoncelli}
Simoncelli, E.P., Portilla, J.:
\newblock Texture characterization via joint statistics of wavelet coefficient
  magnitudes.
\newblock In: ICIP. (1998)

\bibitem{NIPS2015_5633}
Gatys, L., Ecker, A.S., Bethge, M.:
\newblock Texture synthesis using convolutional neural networks.
\newblock In Cortes, C., Lawrence, N.D., Lee, D.D., Sugiyama, M., Garnett, R.,
  eds.: Advances in Neural Information Processing Systems 28.
\newblock Curran Associates, Inc. (2015)  262--270

\bibitem{Johnson2016Perceptual}
Johnson, J., Alahi, A., Fei-Fei, L.:
\newblock Perceptual losses for real-time style transfer and super-resolution.
\newblock In: European Conference on Computer Vision. (2016)

\bibitem{wang2016multimodal}
Wang, X., Oxholm, G., Zhang, D., Wang, Y.F.:
\newblock Multimodal transfer: A hierarchical deep convolutional neural network
  for fast artistic style transfer.
\newblock arXiv preprint arXiv:1612.01895 (2016)

\bibitem{chen2017stylebank}
Chen, D., Yuan, L., Liao, J., Yu, N., Hua, G.:
\newblock Stylebank: An explicit representation for neural image style
  transfer.
\newblock The IEEE Conference on Computer Vision and Pattern Recognition (CVPR)
  (July 2017)

\bibitem{dumoulin2016learned}
Dumoulin, V., Shlens, J., Kudlur, M.:
\newblock A learned representation for artistic style.
\newblock ICLR (2017)

\bibitem{DBLP:journals/corr/UlyanovVL16}
Ulyanov, D., Vedaldi, A., Lempitsky, V.S.:
\newblock Instance normalization: The missing ingredient for fast stylization.
\newblock CoRR \textbf{abs/1607.08022} (2016)

\bibitem{huang2017arbitrary}
Huang, X., Belongie, S.:
\newblock Arbitrary style transfer in real-time with adaptive instance
  normalization.
\newblock In: Proceedings of the IEEE Conference on Computer Vision and Pattern
  Recognition. (2017)  1501--1510

\bibitem{UST}
Li, Y., Fang, C., Yang, J., Wang, Z., Lu, X., Yang, M.H.:
\newblock Universal style transfer via feature transforms.
\newblock arXiv preprint arXiv:1705.08086

\bibitem{chen2016fast}
Chen, T.Q., Schmidt, M.:
\newblock Fast patch-based style transfer of arbitrary style.
\newblock arXiv preprint arXiv:1612.04337 (2016)

\bibitem{Shih2014}
Shih, Y., Paris, S., Barnes, C., Freeman, W.T., Durand, F.:
\newblock {Style transfer for headshot portraits}.
\newblock ACM Transactions on Graphics \textbf{33}(4) (2014)  1--14

\bibitem{Luan2017}
Luan, F., Paris, S., Shechtman, E., Bala, K.:
\newblock {Deep Photo Style Transfer}.
\newblock (2017)

\bibitem{gatys2016controlling}
Gatys, L.A., Ecker, A.S., Bethge, M., Hertzmann, A., Shechtman, E.:
\newblock Controlling perceptual factors in neural style transfer.
\newblock In: The IEEE Conference on Computer Vision and Pattern Recognition
  (CVPR). (July 2017)

\bibitem{li2017demystifying}
Li, Y., Wang, N., Liu, J., Hou, X.:
\newblock Demystifying neural style transfer.
\newblock arXiv preprint arXiv:1701.01036 (2017)

\bibitem{champandard2016semantic}
Champandard, A.J.:
\newblock Semantic style transfer and turning two-bit doodles into fine
  artworks.
\newblock arXiv preprint arXiv:1603.01768 (2016)

\bibitem{jing2017neural}
Jing, Y., Yang, Y., Feng, Z., Ye, J., Song, M.:
\newblock Neural style transfer: A review.
\newblock arXiv preprint arXiv:1705.04058 (2017)

\bibitem{gupta2017characterizing}
Gupta, A., Johnson, J., Alahi, A., Fei-Fei, L.:
\newblock Characterizing and improving stability in neural style transfer

\bibitem{liu1989limited}
Liu, D.C., Nocedal, J.:
\newblock On the limited memory bfgs method for large scale optimization.
\newblock Mathematical programming \textbf{45}(1) (1989)  503--528

\bibitem{bau2017network}
Bau, D., Zhou, B., Khosla, A., Oliva, A., Torralba, A.:
\newblock Network dissection: Quantifying interpretability of deep visual
  representations.
\newblock In: Computer Vision and Pattern Recognition (CVPR), 2017 IEEE
  Conference on, IEEE (2017)  3319--3327

\bibitem{gretton2012kernel}
Gretton, A., Borgwardt, K.M., Rasch, M.J., Sch{\"o}lkopf, B., Smola, A.:
\newblock A kernel two-sample test.
\newblock Journal of Machine Learning Research \textbf{13}(Mar) (2012)
  723--773

\bibitem{DeshpandeCVPR2018}
Deshpande, I., Zhang, Z., Schwing, A.G.:
\newblock {Generative Modeling using the Sliced Wasserstein Distance}.
\newblock In: Proc. CVPR. (2018)

\bibitem{Forsyth2018}
Forsyth, D.
\newblock In: Samples and Populations. Springer International Publishing, Cham
  (2018)  141--157

\bibitem{leongatys}
Gatys, L.:
\newblock Pytorchneuralstyletransfer.
\newblock \url{https://github.com/sunshineatnoon/PytorchWCT} (2017)

\bibitem{abhiskk}
Kadian, A.:
\newblock fast-neural-style.
\newblock \url{https://github.com/abhiskk/fast-neural-style} (2013)

\bibitem{SunshineAtNoon}
Li, X.:
\newblock Universal-style.
\newblock \url{https://github.com/sunshineatnoon/PytorchWCT} (2017)

\end{thebibliography}
}

\section{Additional EC Plots}
We show additional EC plots here, first, we show the rest of 4 layer EC plots where we use probability of boundaries(Pb) AUC as the Coherence measure, then we show EC plots of 5 layers using object coherence $L_m$ (see Sec 3, object coherence) as Coherence measure. 

\subsection{EC plots with AUC for layer 2 to 5}
\begin{figure*}[!htbp]
\centering
    \includegraphics[width=0.95\linewidth]{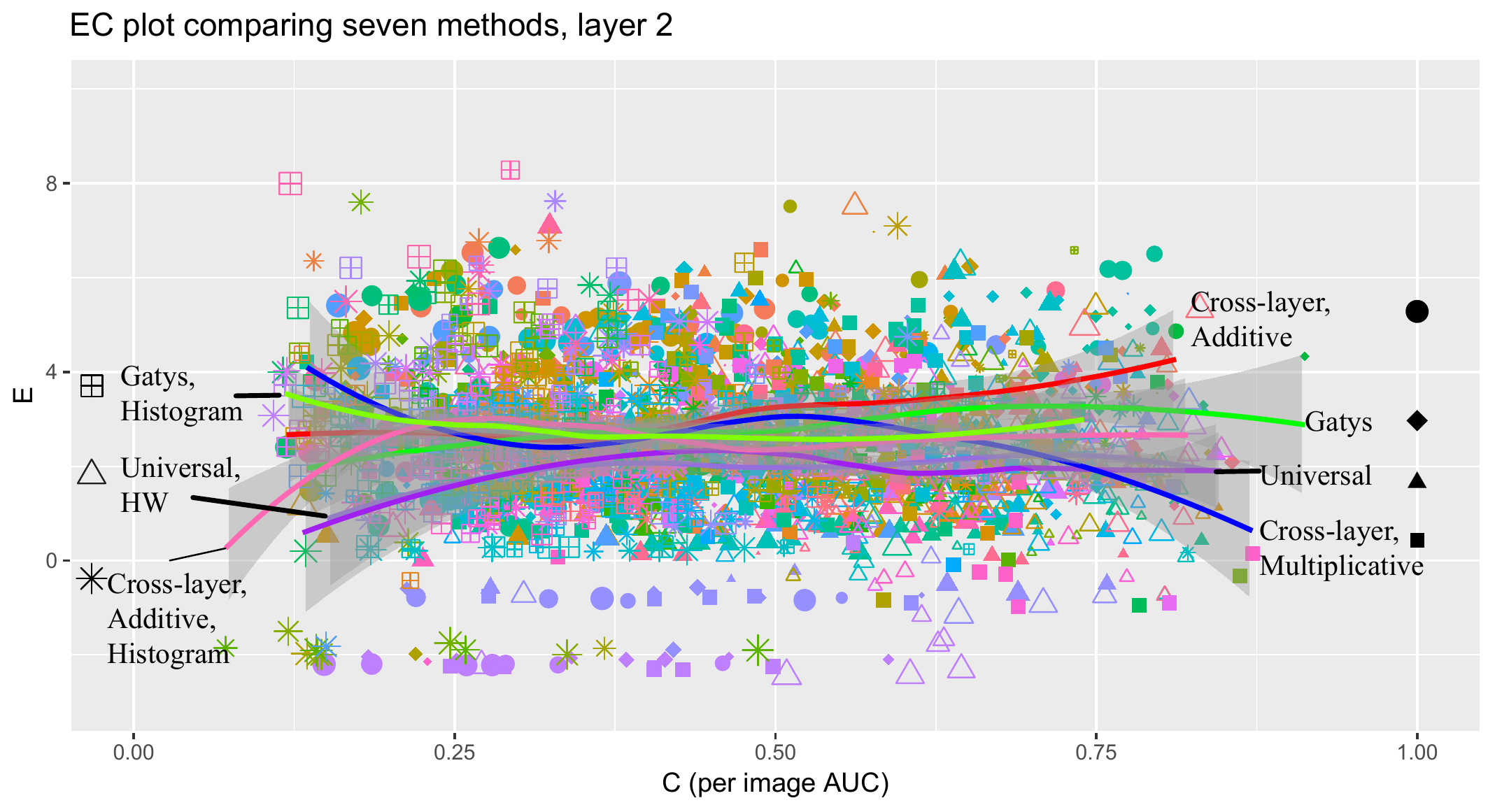}
\caption{\em An EC plot at layer2 compares 7 methods which respectively are Gatys, ACG, Universal style transfer, MCG, Gatys with histogram loss, Universal style transfer with higher weight, ACG with histogram loss.  }
\label{mainweight2}
\vspace{-7mm}
\end{figure*}

\begin{figure*}[!htbp]
\centering
    \includegraphics[width=0.95\linewidth]{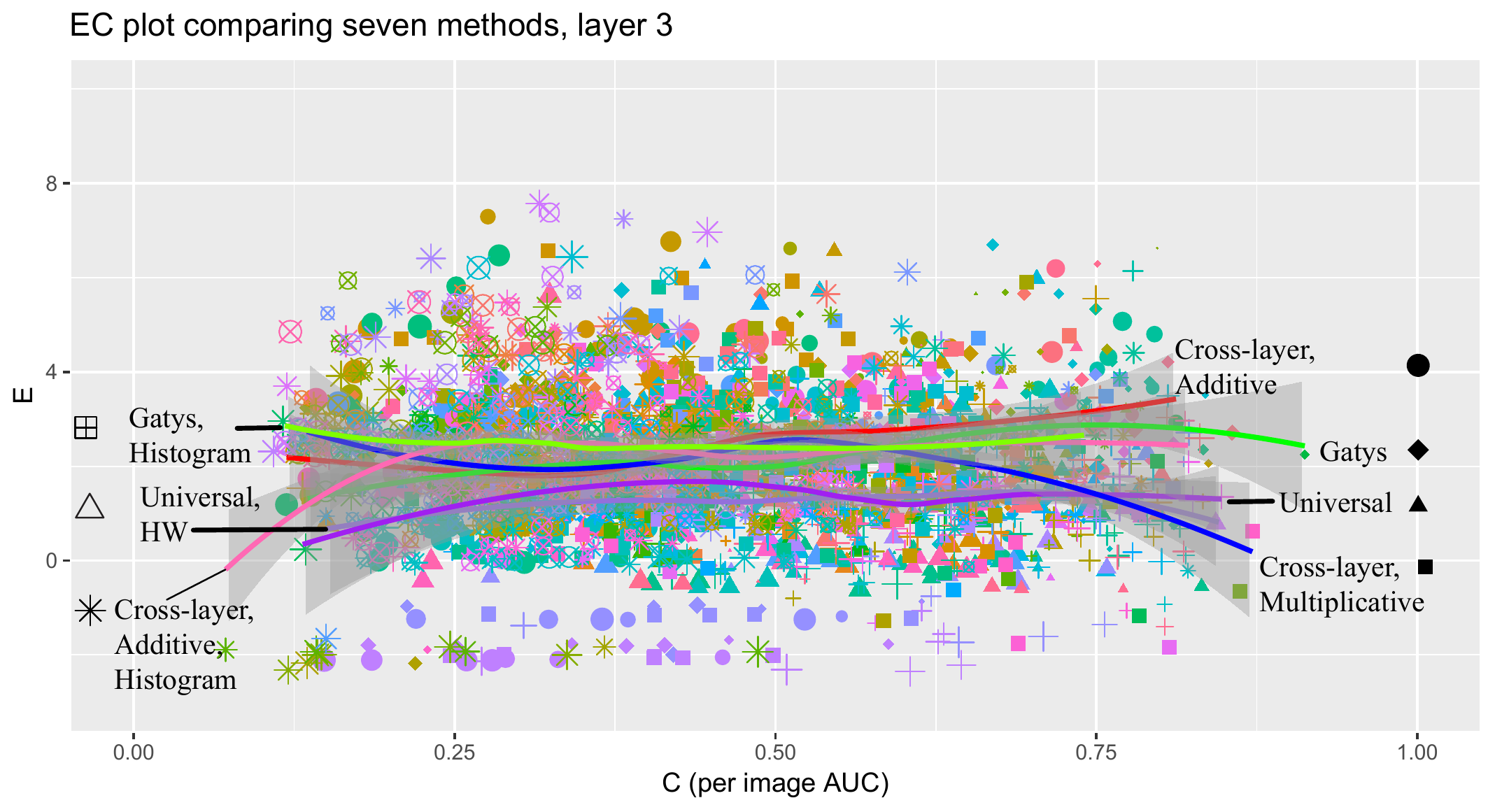}
\caption{\em An EC plot at layer3 compares 7 methods which respectively are Gatys, ACG, Universal style transfer, MCG, Gatys with histogram loss, Universal style transfer with higher weight, ACG with histogram loss.   }
\label{mainweight3}
\vspace{-7mm}
\end{figure*}

\begin{figure*}[!htbp]
\centering
    \includegraphics[width=0.95\linewidth]{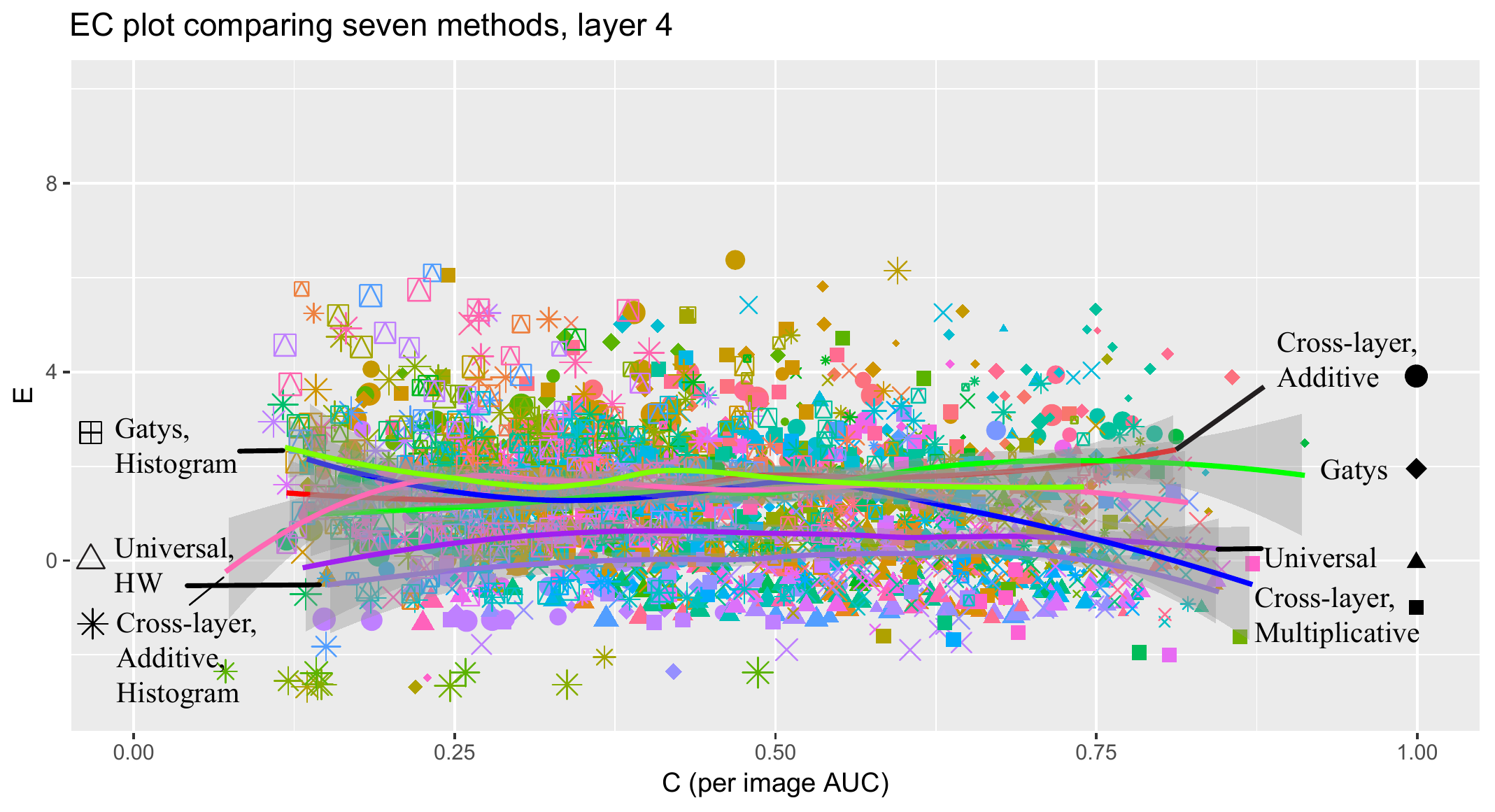}
\caption{\em An EC plot at layer4 compares 7 methods which respectively are Gatys, ACG, Universal style transfer, MCG, Gatys with histogram loss, Universal style transfer with higher weight, ACG with histogram loss.  }
\label{mainweight4}
\vspace{-7mm}
\end{figure*}

\begin{figure*}[!htbp]
\centering
    \includegraphics[width=0.95\linewidth]{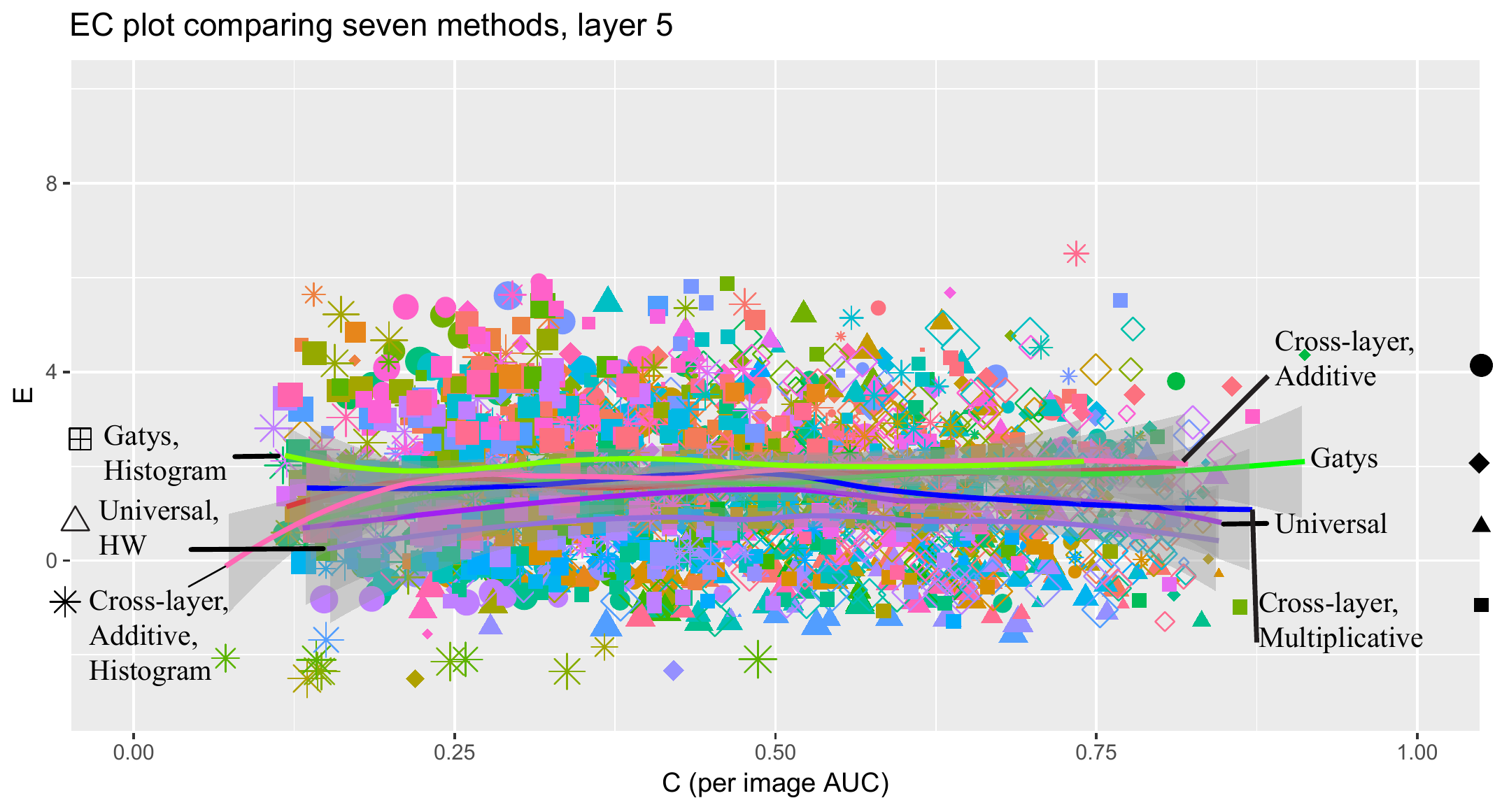}
\caption{\em An EC plot at layer5 compares 7 methods which respectively are Gatys, ACG, Universal style transfer, MCG, Gatys with histogram loss, Universal style transfer with higher weight, ACG with histogram loss.    }
\label{mainweight5}
\vspace{-7mm}
\end{figure*}

Because there is good evidence that the distribution of features within lower feature layers of a CNN 
representation is an effective proxy to capture styles, see [28], we only show  first layer EC plot in our main content.
The rest EC plots are shown in follows.
Figure~\ref{mainweight2},\ref{mainweight3},\ref{mainweight4},\ref{mainweight5} shows EC plots (C as Pb AUC) for layer 2 to 5. the difference between different methods are less marked from layer 1 EC plot.


\FloatBarrier
\FloatBarrier
\subsection{EC plots with $L_m$ ( Objects' Segment Divergence)}
\begin{figure}
\centering
\includegraphics[width=\linewidth]{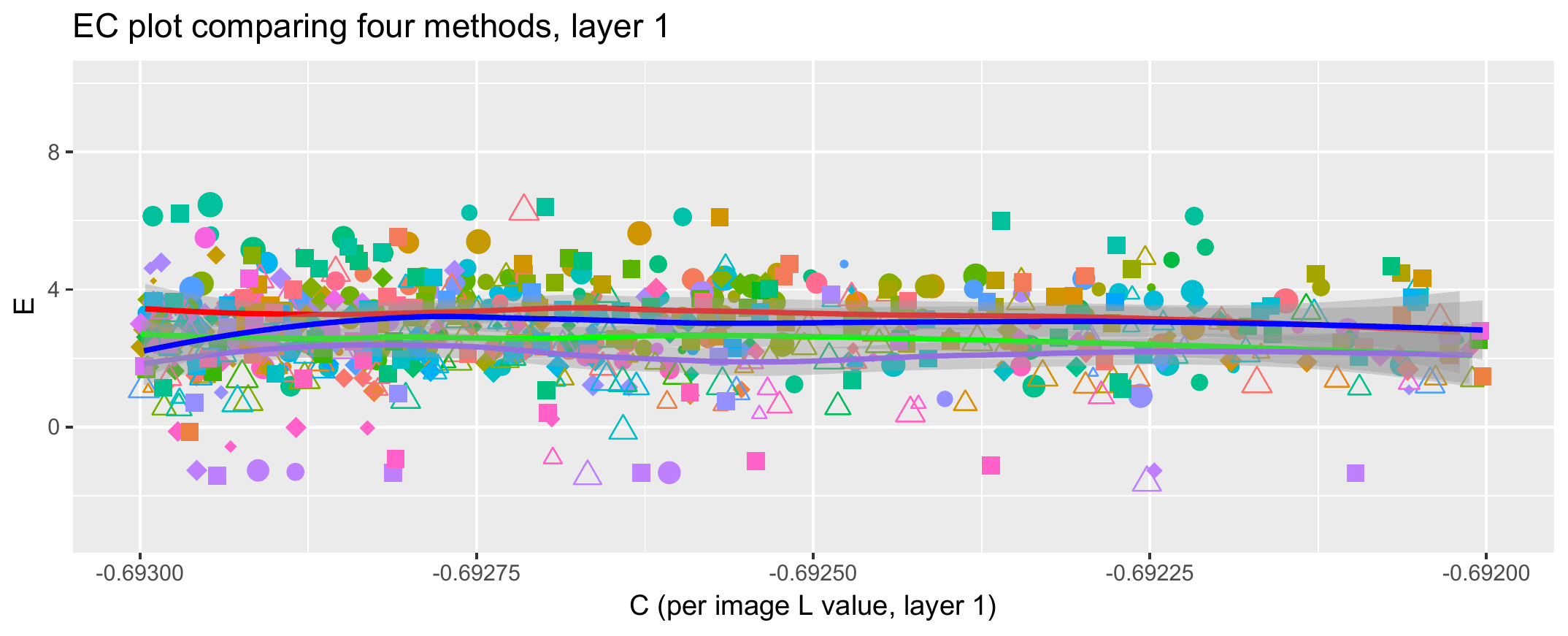}
\caption{EC plot (C as $L_m$) for layer 1, solid lines' color corresponds to previous EC plots; red for ACG, blue for MCG, purple for Universal Style Transfer, and green for Gatys.  }
\label{ECL1}
\end{figure}

\begin{figure}
\centering
\includegraphics[width=\linewidth]{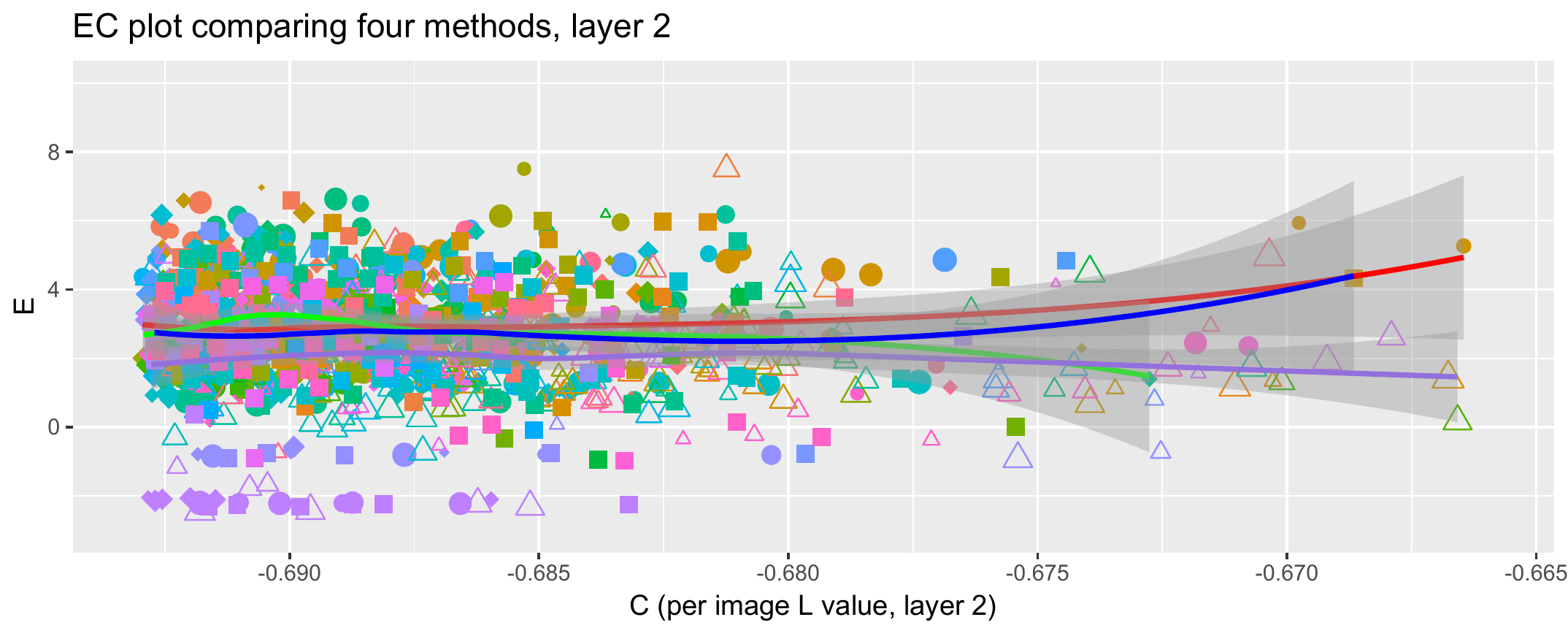}
\caption{EC plot (C as $L_m$) for layer 2, solid lines' color corresponds to previous EC plots; red for ACG, blue for MCG, purple for Universal Style Transfer, and green for Gatys. }
\label{ECL2}
\end{figure}

\begin{figure}
\centering
\includegraphics[width=\linewidth]{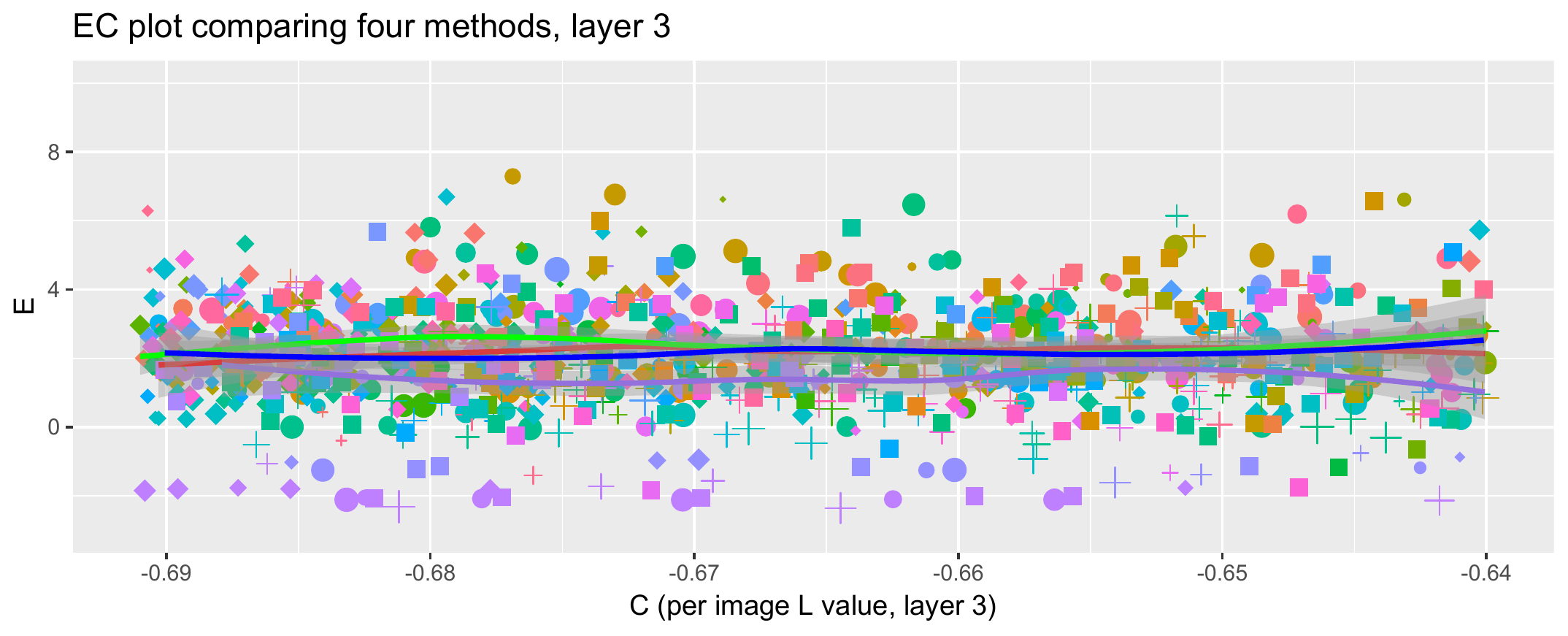}
\caption{EC plot (C as $L_m$) for layer 3, solid lines' color corresponds to previous EC plots; red for ACG, blue for MCG, purple for Universal Style Transfer, and green for Gatys. }
\label{ECL3}
\end{figure}

\begin{figure}
\centering
\includegraphics[width=\linewidth]{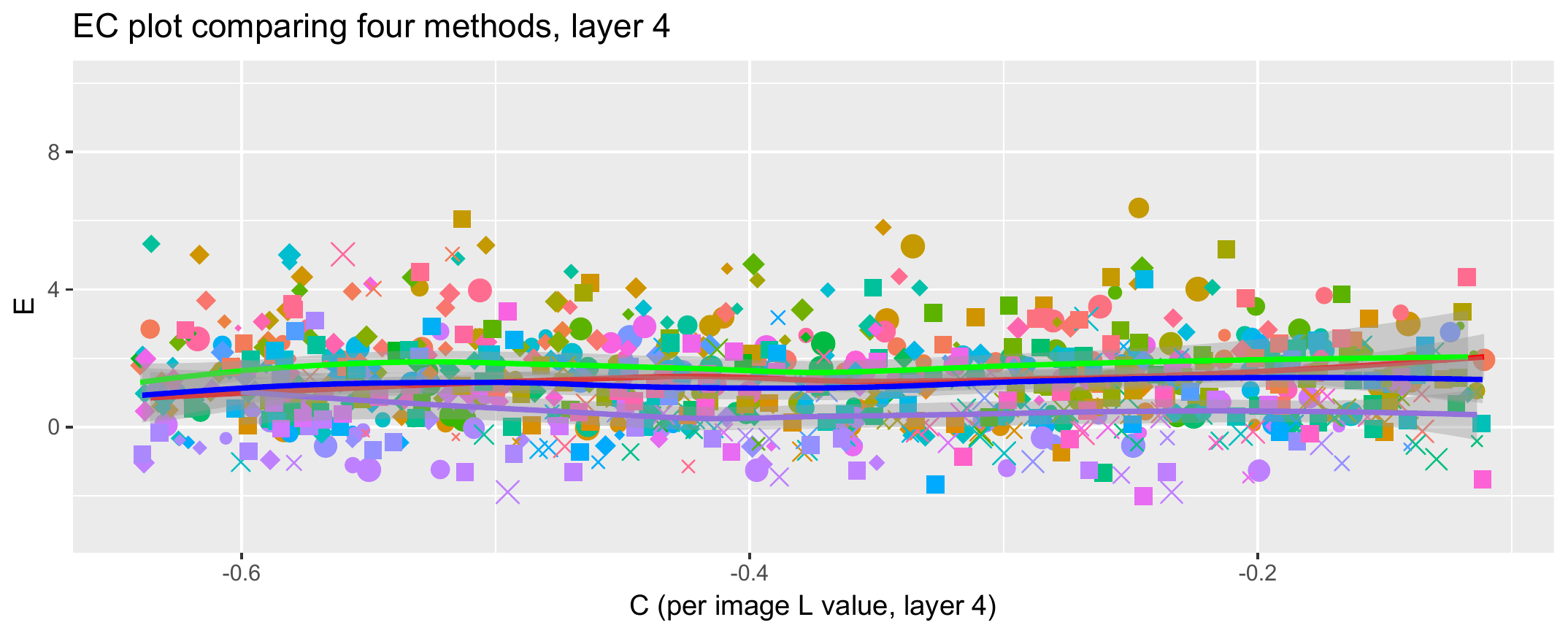}
\caption{EC plot (C as $L_m$) for layer 4, solid lines' color corresponds to previous EC plots; red for ACG, blue for MCG, purple for Universal Style Transfer, and green for Gatys. }
\label{ECL4}
\end{figure}

\begin{figure}
\centering
\includegraphics[width=\linewidth]{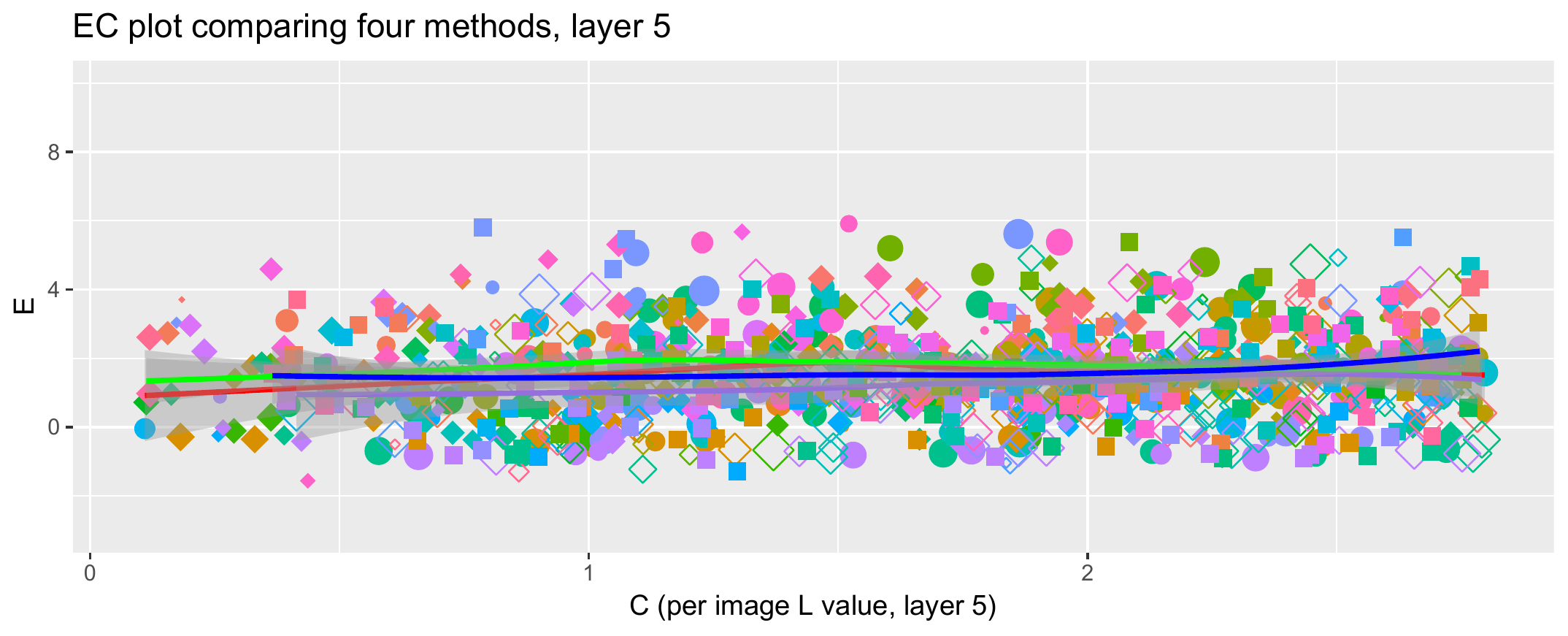}
\caption{EC plot (C as $L_m$) for layer 5, solid lines' color corresponds to previous EC plots; red for ACG, blue for MCG, purple for Universal Style Transfer, and green for Gatys. }
\label{ECL5}
\end{figure}

Figure~\ref{ECL1},\ref{ECL2},\ref{ECL3},\ref{ECL4},\ref{ECL5} shows EC plots (C as $L_m$, see Sec. 3, object coherence) for layer 1 to 5.
 {\bf Because of the extremely compact range of $L_m$, they are not as helpful as AUC.} 

\FloatBarrier

\section{More Qualitative Result Comparison }
\subsection{Quantized EC scatter table}
The Table \ref{sampletable} shows how samples are distributed along the anti-diagonal blocks of 4X4 grid on EC plot. We can see ACG dominates Top-Right Corner.

\begin{table}[]
\centering

\label{tab}
\resizebox{\linewidth}{!}{%
\begin{tabular}{|
>{\columncolor[HTML]{EFEFEF}}c |c|c|c|c|c|}
\hline
Anti-Diagonal Position  & \multicolumn{1}{l|}{\cellcolor[HTML]{E1E1E1}ACG(Ours)} & \multicolumn{1}{l|}{\cellcolor[HTML]{E1E1E1}Gatys} & \cellcolor[HTML]{E1E1E1}ACG +Hist & \cellcolor[HTML]{E1E1E1}Gatys + Hist & \cellcolor[HTML]{E1E1E1}UST \\ \hline
Top-Right Corner (IV)        & \textbf{52}                                                & 20                                                 & 5                                            & 1                                          & 19                                      \\ \hline
Second from Top-Right (III)   & 26                                                         & 45                                                 & 38                                           & 32                                         & 42                                      \\ \hline
Second from Bottom-Left (II) & 20                                                         & 15                                                 & 25                                           & 44                                         & 8                                       \\ \hline
Bottom-Left Corner (I)      & 14                                                         & 17                                                 & 27                                           & 40                                         & \textbf{5}                              \\ \hline
\end{tabular}}
\vspace{3mm}
\caption{{\bf Quantized EC Scatter plots.} After combining all 1500 samples of \textit{main set} (5 methods and 300 stylized images in each), we divide these samples in a 4x4 grid and report number of samples in anti-diagonal locations. Top row indicates best stylized images (High E \& High C) and bottom row indicates poor style transfer (Low E \& Low C ). }
\label{sampletable}
\vspace{-7mm}
\end{table}

\subsection{ Landscape images mediods from 4 blocks of EC plot }
\begin{figure*}[!htbp]
    \centering
    \includegraphics[width=\linewidth]{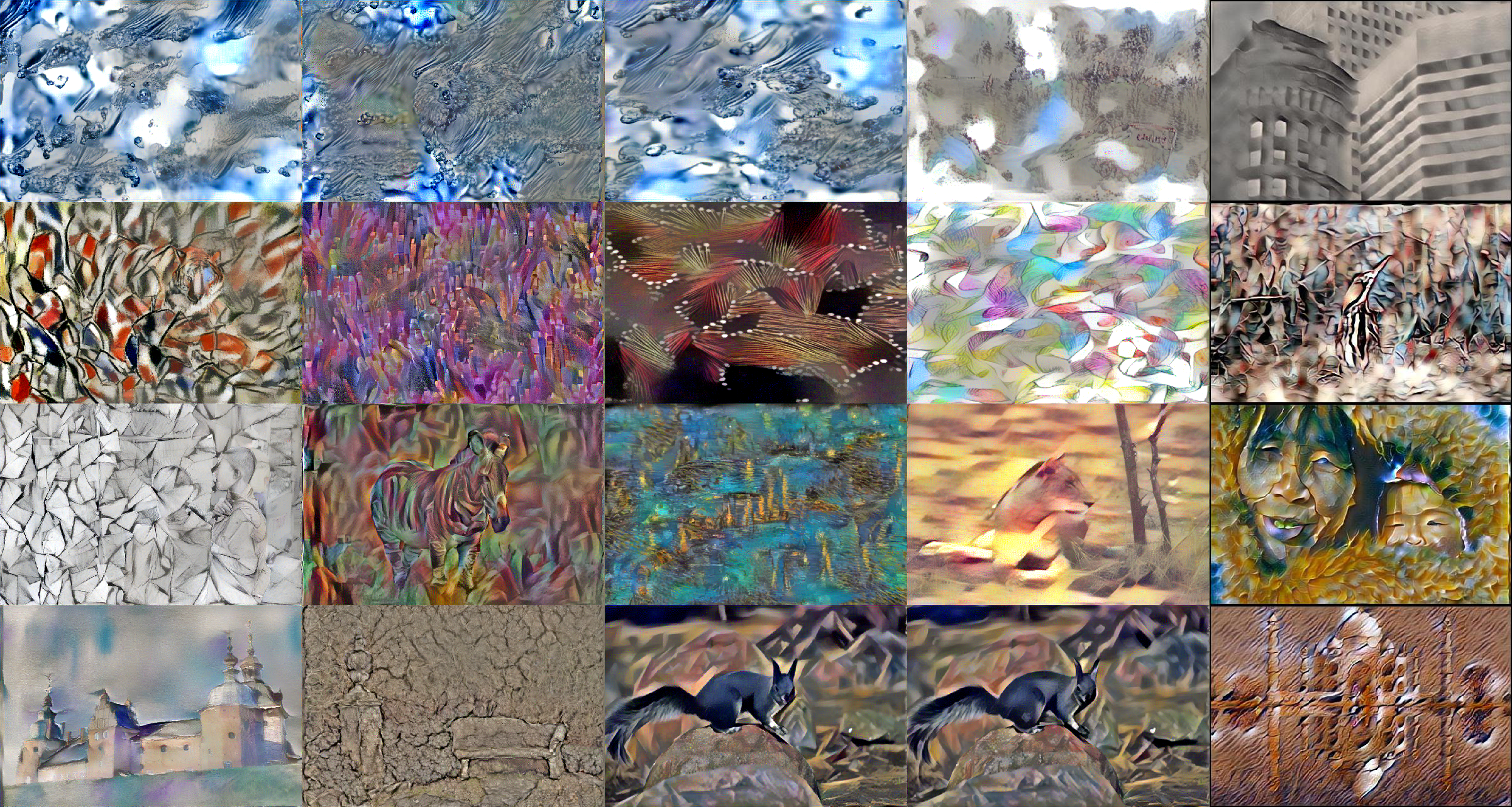}
    \caption{
Mediods from different blocks in the EC plot for five style transfer methods reveal the semantics of the EC statistics.
We divided the EC plot into a 4x4 grid, using quantiles of values of E and C over all methods.
Each {\bf row} corresponds to a different block along the diagonal in this grid, ranging from bottom left grid box
(with lowest E and C, so worst, {\bf top row}) to top right (with highest E and C, so best, {\bf bottom row}).
Each {\bf column} represents a method, in order from left to right: {\bf ACG}, {\bf Gatys}, {\bf ACG+Histogram}, {\bf Gatys+Histogram}, {\bf UST}.
    }  
    \label{fig:qual_land}
    \vspace{-5mm}
    \end{figure*}

Figure \ref{fig:qual_land} using the same block division in Figure 1 of our paper, we show more qualitative results for landscape images (of dimension 481-by-321 pixels ).

\subsection{Easy and Hard Styles}
\begin{figure*}[!htbp]
    \centering
    \includegraphics[width=\linewidth]{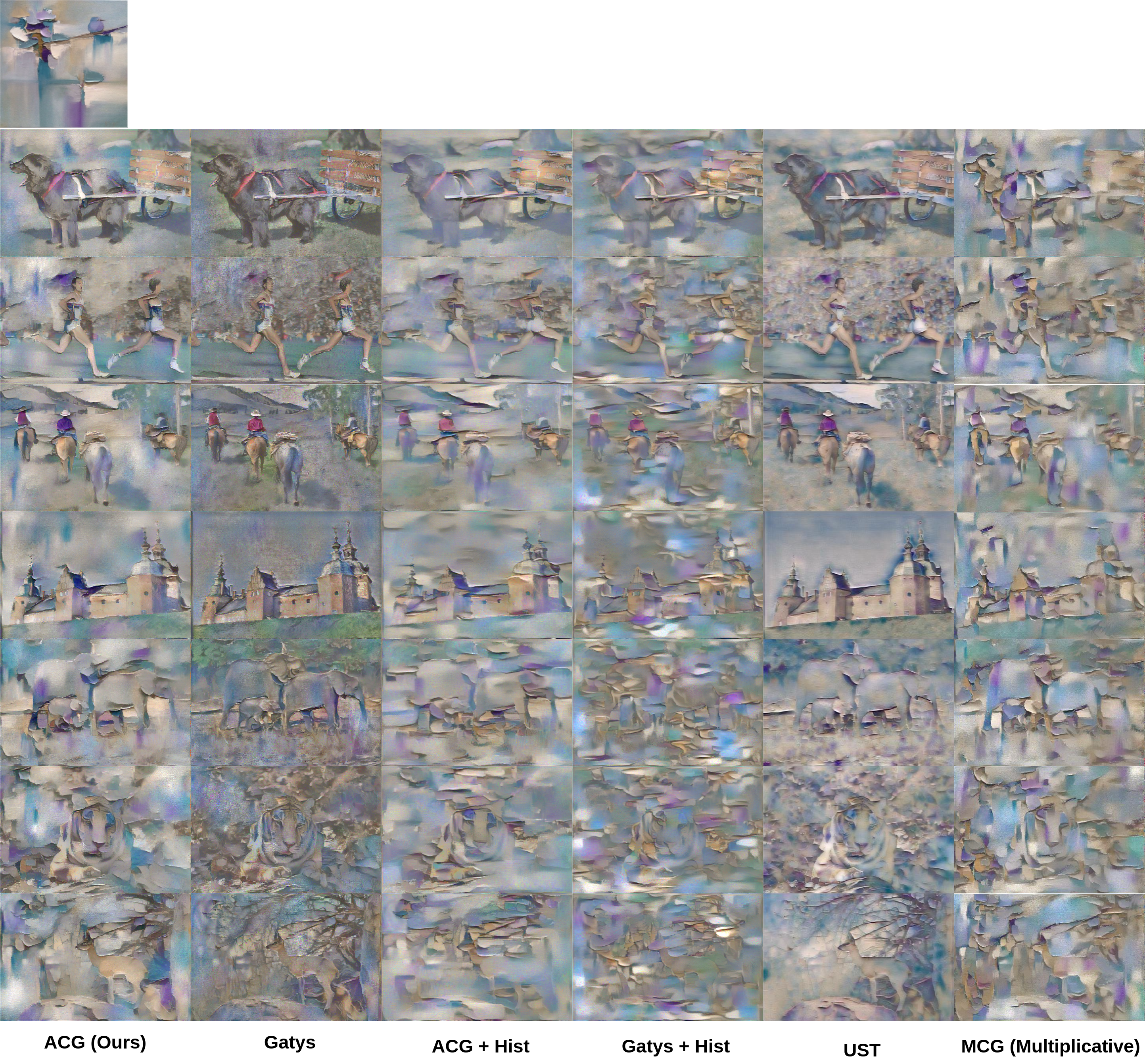}
    \caption{For given easy style, which is easy style because most methods can gives high E value, the images from left column to right column respectively are generated  by {\bf ACG, Gatys, ACG with histogram loss, Gatys with histogram loss, UST(universal style transfer), and MCG}.  The rows represent different contents. 
    }  
    \label{fig:qual_best}
    \vspace{-5mm}
    \end{figure*}

\begin{figure*}[!htbp]
    \centering
    \includegraphics[width=\linewidth]{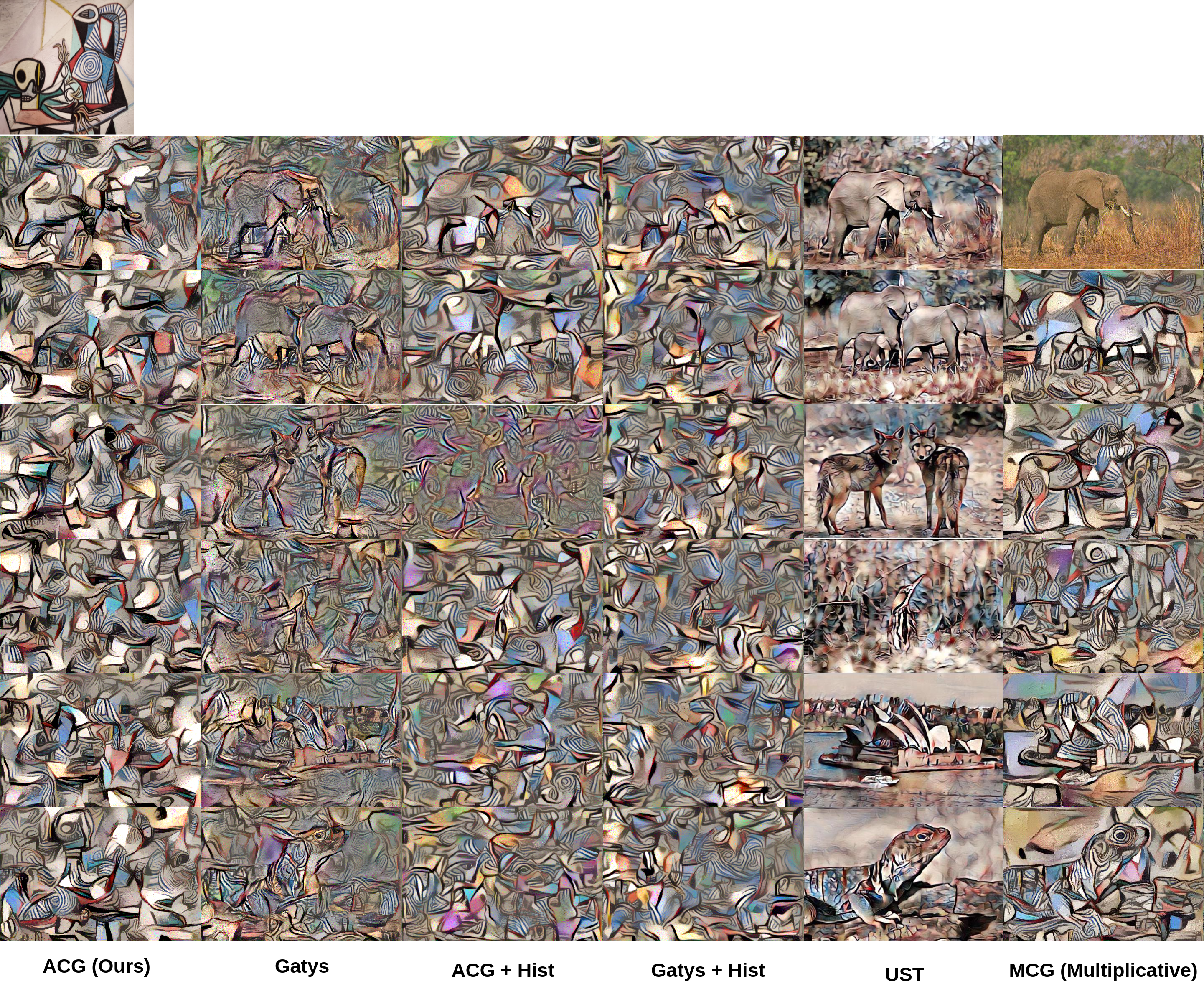}
    \caption{ Based on a given hard style,  which is hard  because no methods can gives high E value, the images from left column to right column respectively generated  by {\bf ACG, Gatys, ACG with histogram loss, Gatys with histogram loss, UST(universal style transfer), and MCG}.  The rows represent different contents. 
    }  
    \label{fig:qual_diff}
    \vspace{-5mm}
    \end{figure*}

Figure \ref{fig:qual_best} and Figure \ref{fig:qual_diff}  show how images actually looks in our samples for given two extremely styles (easy and hard). The easy style and hard style have extreme high  E and  extreme low  E respectively.
\FloatBarrier
\FloatBarrier

\section{Selected 50 Styles}
\begin{figure*}[bt]
\centering
  \includegraphics[width=0.6\linewidth]{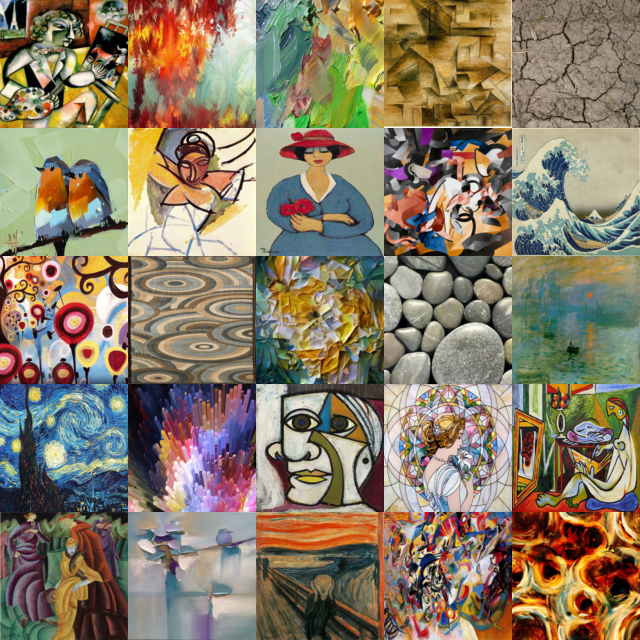}
\caption{\em The first group of 50 styles.     }
\label{50styles1}
\vspace{-3mm}
\end{figure*}

\begin{figure*}[bt]
\centering
  \includegraphics[width=0.6\linewidth]{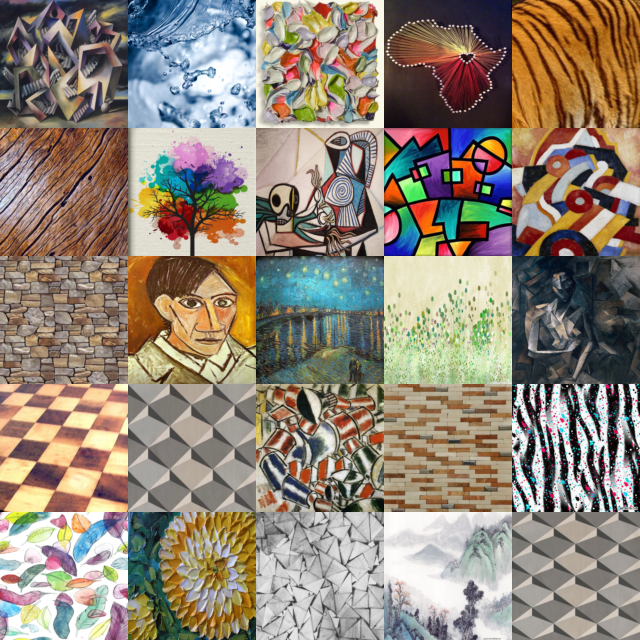}
\caption{\em The second group of 50 styles.     }
\label{50styles2}
\vspace{-3mm}
\end{figure*}

Figure \ref{50styles1} and Figure \ref{50styles2} display our 50 style images. Except the Universal style transfer, all other methods synthesize image from Gaussian noise with LBFGS optimizer. The content images and style images are resized to same width of 512 as the input for style transfers. 

\FloatBarrier

\section{Construction of Affine Maps for Symmetry Groups}
This difference in symmetry groups is important.  Risser argues that
the symmetries of gram matrices in Gatys' method could lead to
unstable reconstructions; they control this effect using feature
histograms.  What causes the effect is that the symmetry rescales features while shifting the mean.  
For the cross-layer loss, the symmetry cannot rescale, and cannot shift the mean.  In turn, the instability
identified in that paper does not apply to the cross-layer gram matrix and our results could not be improved by adopting
a histogram loss.

Write $\vect{x}_{i}$, (resp $\vect{y}_i$ for the feature vector at the $i$'th location (of $N$ in total)
in the first (resp second) layer.  Write $\matx{X}^T=\left[\vect{x}_{1}, \ldots,
\vect{x}_{N}\right]$, etc.   

{\bf Symmetries of the first layer:} Now assume that the first layer has been normalized to zero mean and
unit covariance.  There is no loss of generality, because the whitening transform
can be written into the expression for the group. Write ${\cal
  G}(\matx{W})=(1/N)\matx{W}^T\matx{W}$ for the operator that forms
the within layer gram matrix. We have ${\cal G}(\matx{X})=\matx{I}$.   
Now consider an affine action on layer 1, mapping $\matx{X}_1$ to $\matx{X}_1^*=\matx{X}_1 \matx{A}+\vect{1}\vect{b}^T$; then for this to be a symmetry, we must have
$G(\matx{X}_1^*)= \matx{A}\matx{A}^T+\vect{b}\vect{b}^T=\matx{I}$.  In
turn, the symmetry group can be constructed by: choose $\vect{b}$
which does not have unit length; factor
$N(\matx{I}-\vect{b}\vect{b}^T)$ to obtain $\matx{A}(\vect{b})$ (for
example, by using a cholesky transformation); then any element of the
group is a pair $\left(\vect{b}, \matx{A}(\vect{b})\matx{U}\right)$
where $\matx{U}$ is orthonormal.  Note that factoring will fail for
$\vect{b}$ a unit vector, whence the restriction. 

{\bf The second layer:}   We will assume that the map between layers
of features is linear.  This assumption is not true in practice, but major
differences between symmetries observed under these conditions likely
result in differences when the map is linear.  We can analyze for two
cases: first, all units in the map observe only one input feature
vector (i.e. 1x1 convolutions; the {\em point sample} case); second, spatial homogeneity in the
layers.

{\bf The point sample case:} Assume that every unit in the map
observes only one input feature from the previous layer (1x1
convolutions).   We have $\matx{Y}=\matx{X}\matx{M}+\vect{1}\vect{n}^T$, because the map 
between layers is linear. Now consider the effect on the second layer.
We have ${\cal G}(\matx{Y})=\matx{M}\matx{M}^T+\vect{n}\vect{n}^T$.
Choose some symmetry group element for the first layer,
$\left(\vect{b}, \matx{A}\right)$.  The gram matrix for the second
layer becomes ${\cal G}(\matx{Y}^*)$, where
$\matx{Y}^*=(\matx{X} \matx{A}+\vect{1}\vect{b}^T)\matx{M}^T+\vect{1}\vect{n}^T.$
Recalling that $\matx{A}\matx{A}^T+\vect{b}\vect{b}^T=\matx{I}$ and
$\matx{X}^T\vect{1}=0$, we have 
\[
{\cal
  G}(\matx{Y}^*)=\matx{M}\matx{M}^T+\vect{n}\vect{n}^T+\vect{n}\vect{b}^T\matx{M}^T+\matx{M}\vect{b}\vect{n}^T
\]
so that ${\cal G}(\matx{X}_2^*)={\cal G}(\matx{X}_2)$ if
$\matx{M}\vect{b}=0$.  This is relatively easy to achieve with
$\vect{b}\neq 0$.

{\bf Spatial homogeneity:} Now assume the map between layers has
convolutions with maximum support $r \times r$.  Write $u$ for an
index that runs over the whole feature map, and $\psi(\vect{x}_u)$ for
a stacking operator that scans the convolutional support in fixed
order and stacks the resulting features. For example, given a 3x3
convolution and indexing in 2D, we might have
\[
\psi(\vect{x}_{22})=\left(\begin{array}{c}\vect{x}_{11}\\
\vect{x}_{12}\\
\ldots\\
\vect{x}_{33}
\end{array}\right)
\]

In this case, there is some $\matx{M}$, $\vect{n}$ so that 
$\vect{y}_u=\matx{M}\psi(\vect{x}_u)+\vect{n}$.  We ignore the
effects of edges to simplify notation (though this argument may go
through if edges are taken into account).  Then there is some
$\matx{M}$, $\vect{n}$ so we can write 
\[
{\cal G}(\matx{Y})=(1/N) \sum_u
\matx{M}\psi(\vect{x}_u)\psi(\vect{x}_u)^T\matx{M}^T+\vect{n}\vect{n}^T
\]
Now assume further that
layer 1 has the following (quite restrictive) spatial homogeneity
property: for pairs of feature vectors within the layer $\vect{x}_{i,j}$, $\vect{x}_{i+\delta,
  j+\delta}$ with $\mid \! \delta\!\mid \leq r$ (ie within a convolution window of one
another), we have $\expect{\vect{x}_{i,j}\vect{x}_{i+\delta,
    j+\delta}}=\matx{I}$.  This assumption is consistent with image
autocorrelation functions (which fall off fairly slowly), but is still
strong. Write $\phi$ for an operator
that stacks $r \times r$ copies of its argument as appropriate, so
\[\phi(\matx{I})=\left(\begin{array}{ccc}
\matx{I}&\ldots&\matx{I}\\
\ldots &\ldots \ldots \\
\matx{I}&\ldots&\matx{I}
\end{array}\right).
\]
Then
$G(\matx{Y})=\matx{M}\phi(\matx{I})\matx{M}^T+\vect{n}\vect{n}^T$.
If there is some affine action on layer 1, we have
$G(\matx{Y}^*)=\matx{M}\left(\psi(\matx{A})\phi(\matx{I})\psi(\matx{A}^T)+\psi(\vect{b})\psi(\vect{b}^T)\right)\matx{M}^T+\vect{n}\vect{n}^T$,
where we have overloaded $\psi$ in the natural way.  Now if
$\matx{M}\psi(\vect{b})=0$ and $\matx{A}\matx{A}^T+\vect{b}\vect{b}^T=\matx{I}$, ${\cal
  G}(\matx{Y}^*)={\cal G}(\matx{Y})$.

{\bf The cross-layer gram matrix:}  Symmetries of the cross-layer gram matrix are very different.  Write
${\cal G}(\matx{X}, \matx{Y})=(1/N) \matx{X}^T\matx{Y}$ for
the cross layer gram matrix.  

{\bf Cross-layer, point sample case:} Here (recalling $\matx{X}^T\vect{1}=0$)we have ${\cal G}(\matx{X},
\matx{Y})=\matx{M}^T$.    Now choose some symmetry group element for the first layer,
$\left(\matx{A}, \vect{b}\right)$.  The cross-layer gram matrix
becomes 
\begin{eqnarray*}
{\cal G}(\matx{X}^*, \matx{Y}^*)&=&(1/N) (\matx{A}
\matx{X}^T+\vect{b}\vect{1}^T)
\left[(\matx{X}
  \matx{A}^T+\vect{1}\vect{b}^T)\matx{M}^T+\vect{1}\vect{n}^T\right]\\
&=&\matx{M}^T+\vect{b}\vect{n}^T
\end{eqnarray*}
(recalling that $\matx{A}\matx{A}^T+\vect{b}\vect{b}^T=\matx{I}$ and
$\matx{X}^T\vect{1}=0$).  But this means that the symmetry requires
$\vect{b}=\vect{0}$; in turn, we must have $\matx{A}\matx{A}^T=\matx{I}$.

{\bf Cross-layer, homogeneous case:} We have
\[
{\cal G}(\matx{X}, \matx{Y})=(1/N)\sum_u \vect{x}_u\left[
  \psi(\vect{x}_u)^T\matx{M}^T+\vect{n}^T\right]=\matx{M}^T.\]
Now choose some symmetry group element for the first layer,
$\left(\matx{A}, \vect{b}\right)$.  The cross-layer gram matrix
becomes 
\begin{eqnarray*}
{\cal G}(\matx{X}^*, \matx{Y}^*)&=&(1/N)\sum_u \left(\matx{A} \vect{x}_u+\vect{b}\right)\left[\left(\psi(\vect{x}_u)^T\psi(\matx{A}^T)+\psi(\vect{b})\right)\matx{M}^T+\vect{n}^T\right]\\
&=&\matx{M}^T+\vect{b}\vect{n}^T
\end{eqnarray*}
(recalling the spatial homogeneity assumption, that $\matx{A}\matx{A}^T+\vect{b}\vect{b}^T=\matx{I}$ and
$\matx{X}_1^T\vect{1}=0$).  But this means that the symmetry requires
$\vect{b}=\vect{0}$; in turn, we must have
$\matx{A}\matx{A}^T=\matx{I}$.

\section{Texture Synthesis of Gatys and ACG}

From texture synthesis result, we can easily see ACG preserve better long scale structure of style than Gatys, e.g. the ACG has larger and ordered blocks than Gatys in the third  column.
\begin{figure}
\centering
\includegraphics[width=\linewidth]{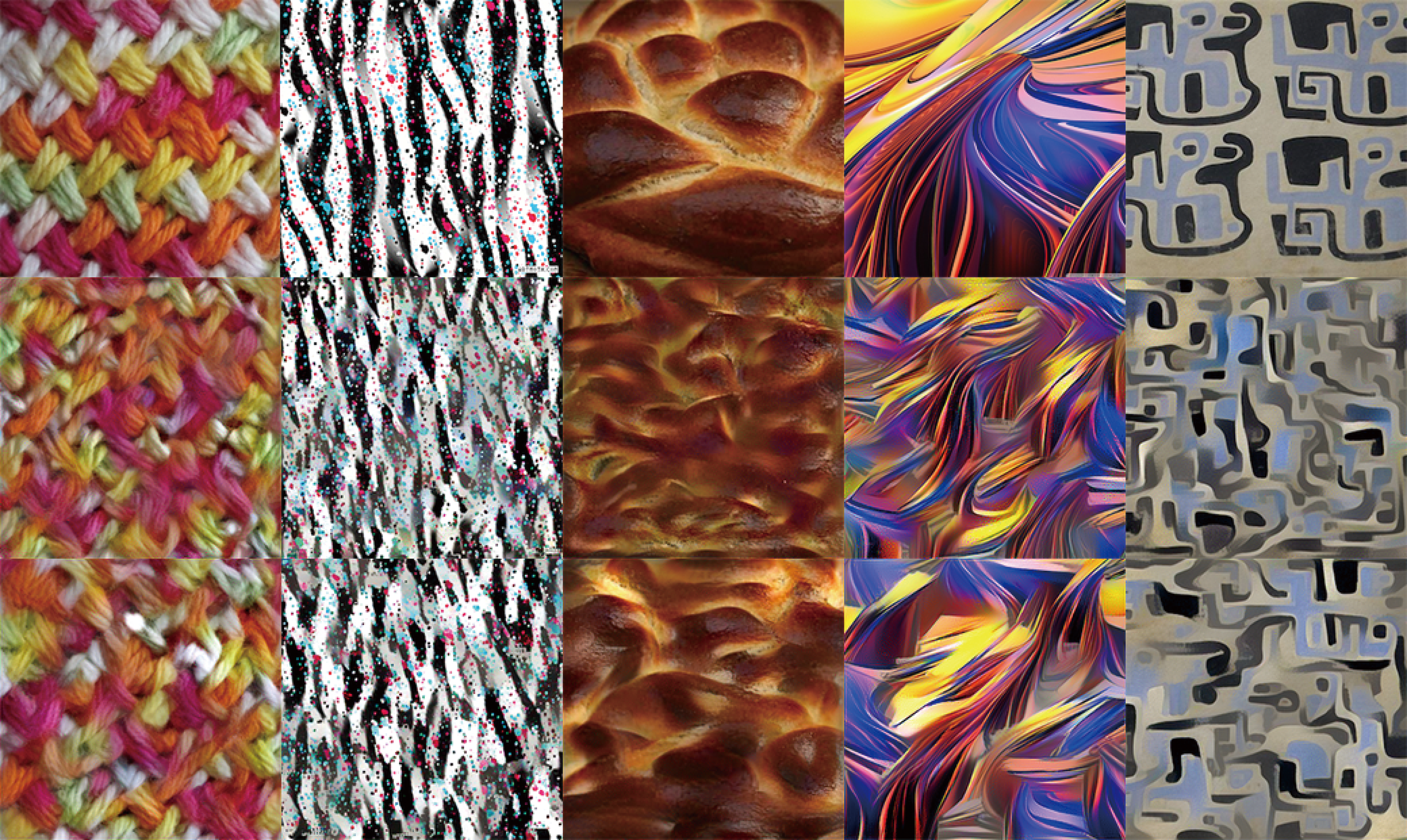}
\caption{\em Qualitatively comparing texture synthesis results. \textbf{First row:}original textures; \textbf{second:} Gatys; \textbf{thrid:} ACG; }
\end{figure}

\section{Distribution of eigenvalues}

\begin{figure*}[h!]
\centering
  \includegraphics[width=0.7\linewidth]{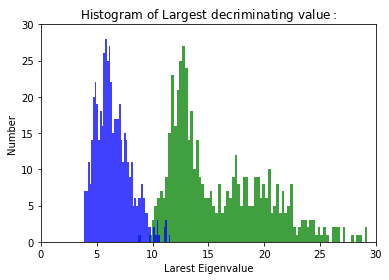}
\caption{\em The histogram plot of eigenvalue of Gatys(blue) and ACG(green) shows a log-normal distribution over the samples.     }
\label{lognormal}
\vspace{-3mm}
\end{figure*}

Figure \ref{lognormal} shows that the distribution of largest generalized eigenvalue over the samples behaves like a log-normal distribution rather than normal distribution.
\FloatBarrier

\end{document}